\documentclass[APA,LATO2COL]{WileyNJD-v2}

\usepackage[utf8]{inputenc}
\usepackage{subfig}
\usepackage{soul}

\usepackage{graphicx}
\usepackage{amsfonts}
\usepackage{bm}
\usepackage{algorithm}
\usepackage{algpseudocode}
\usepackage{caption}
\usepackage{amssymb}

\DeclareUnicodeCharacter{00B4}{'}

\algnewcommand\algorithmicforeach{\textbf{for each}}
\algdef{S}[FOR]{ForEach}[1]{\algorithmicforeach\ #1\ \algorithmicdo}

\articletype{Article Type}

\received{1 September, 2023}
\revised{1 September, 2023}
\accepted{1 September, 2023}

\begin{document}

\title{Open Source Robot Localization for Non-Planar Environments}

\author[1]{Francisco Martín Rico}
\author[1]{José Miguel Guerrero Hernández}
\author[1]{Rodrigo Pérez-Rodríguez}
\author[2]{Juan Diego Peña-Narvaez}
\author[2]{Alberto García Gómez-Jacinto}

\authormark{Francisco Martín Rico \textsc{et al}}

\address[1]{\orgname{Intelligent Robotics Lab. Universidad Rey Juan Carlos, Fuenlabrada}, \orgaddress{\state{Madrid}, \country{Spain}}}

\address[2]{\orgname{International Doctoral School, Rey Juan Carlos University, Móstoles, Spain}, \orgaddress{\state{Madrid}, \country{Spain}}}

\corres{*Francisco Martín Rico, Intelligent Robotics Lab. Universidad Rey Juan Carlos, Fuenlabrada, Madrid. \email{francisco.rico@urjc.es}}

\abstract[Abstract]{

The operational environments in which a mobile robot executes its missions often exhibit non-flat terrain characteristics, encompassing outdoor and indoor settings featuring ramps and slopes. In such scenarios, the conventional methodologies employed for localization encounter novel challenges and limitations. This study delineates a localization framework incorporating ground elevation and incline considerations, deviating from traditional 2D localization paradigms that may falter in such contexts. In our proposed approach, the map encompasses elevation and spatial occupancy information, employing Gridmaps and Octomaps. At the same time, the perception model is designed to accommodate the robot's inclined orientation and the potential presence of ground as an obstacle, besides usual structural and dynamic obstacles. We provide an implementation of our approach fully working with Nav2, ready to replace the baseline AMCL approach when the robot is in non-planar environments. Our methodology was rigorously tested in both simulated environments and through practical application on actual robots, including the Tiago and Summit XL models, across various settings ranging from indoor and outdoor to flat and uneven terrains. Demonstrating exceptional precision, our approach yielded error margins below 10 centimeters and 0.05 radians in indoor settings and less than 1.0 meters in extensive outdoor routes. While our results exhibit a slight improvement over AMCL in indoor environments, the enhancement in performance is significantly more pronounced when compared to 3D SLAM algorithms. This underscores the considerable robustness and efficiency of our approach, positioning it as an effective strategy for mobile robots tasked with navigating expansive and intricate indoor/outdoor environments.}


\jnlcitation{\cname{%
\author{F. Martín-Rico}, 
\author{J.M. Guerrero-Hernández}, 
\author{R. Pérez-Rodríguez},
\author{J.D. Peña-Narváez}, and 
\author{A. Gómez-Jacinto}} (\cyear{2023}), 
\ctitle{Open Source Robot Localization for Non-Planar Environments}, \cjournal{Journal of Field Robotics.}, \cvol{2023;XX:N--M}.}

\maketitle

\section{INTRODUCTION}\label{sec:intro}

Mobile robots have proliferated globally, extending their presence beyond conventional office settings. Autonomous vehicles have emerged as a prominent research frontier, necessitating navigation through diverse terrains encompassing bridges, tunnels, and slopes. Various industries, including agriculture, space exploration, outdoor surveillance, and delivery, are progressively integrating robotic systems into their operations. However, in many of these environments, no standardized approach exists, or existing standard solutions are adapted for relatively simple terrains.

When discussing contemporary robotics standards, the Robot Operating System~\cite{doi:10.1126/scirobotics.abm6074} (ROS) takes center stage, and within the navigation domain, Nav2~\cite{macenski2020marathon2} stands as a pivotal reference framework. Nowadays Nav2 orchestrates the movement of countless robots worldwide, spanning hundreds of companies, organizations, and academic institutions. Despite its widespread adoption, Nav2 uses a 2D perspective, assuming the robot's position as $(x, y, \theta)$ while encoding the map in a 2D occupancy grid. This approach primarily caters to flat environments, failing to address uneven terrains, which are predominant in outdoor environments.

Our motivation is to extend the capabilities of Nav2, transcending this inherent limitation and enabling its application in scenarios characterized by terrain irregularities. This endeavor presents multifaceted challenges. In this paper, our primary focus lies on two crucial aspects: environment encoding and robot localization, both of which serve as the foundational pillars for future navigation in non-planar environments. Our objective extends beyond merely achieving functionality; we aim for robustness, precision, and reliability while seamlessly integrating these enhancements into the existing complex framework of Nav2. This seamless integration ensures a smooth transition, if necessary, without disrupting the established development model and ongoing dynamics of the Nav2 Open Source community.

\begin{figure}[tb]
  \centering
  \includegraphics[width=1\linewidth]{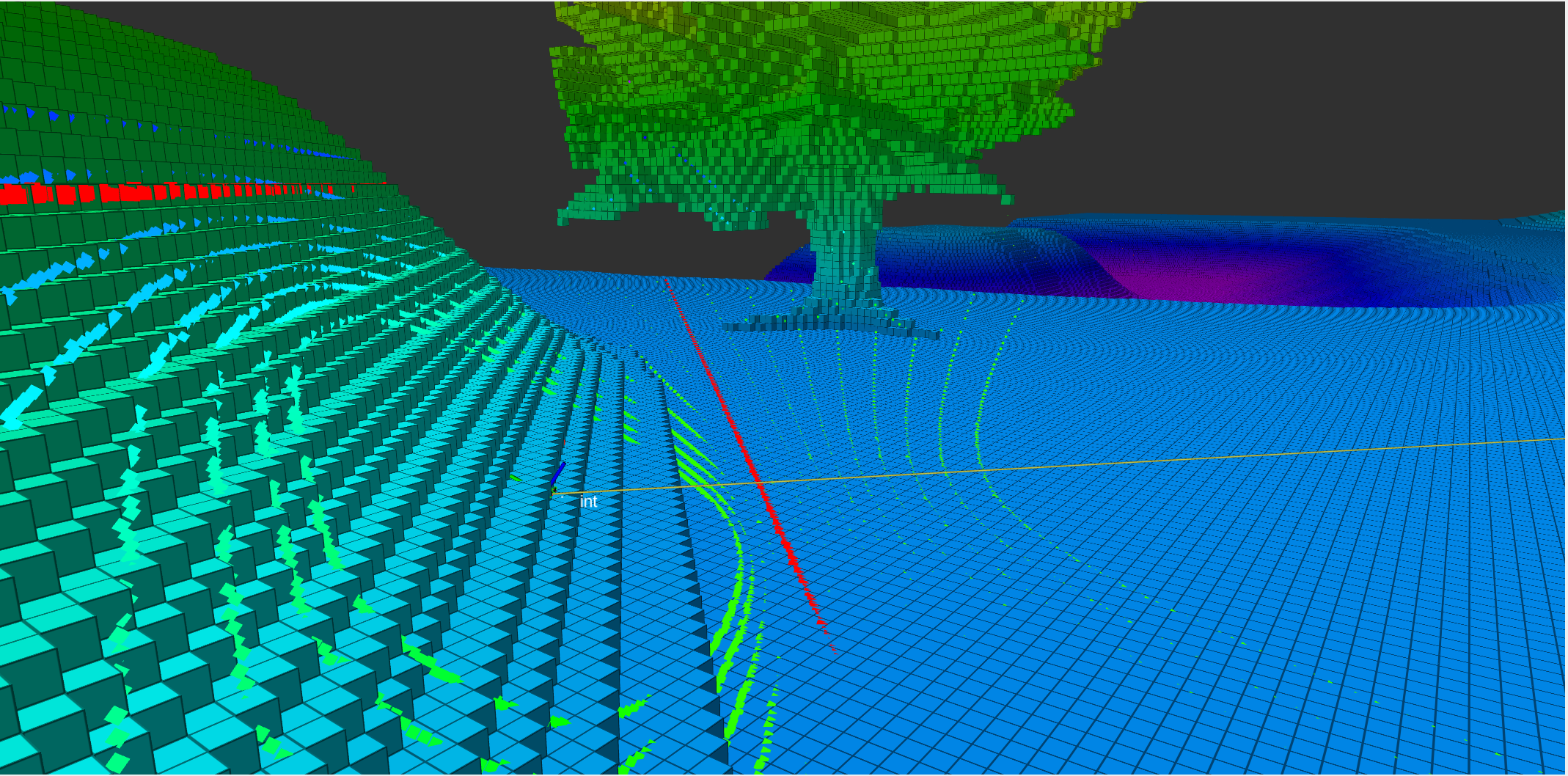}
  \includegraphics[width=1\linewidth]{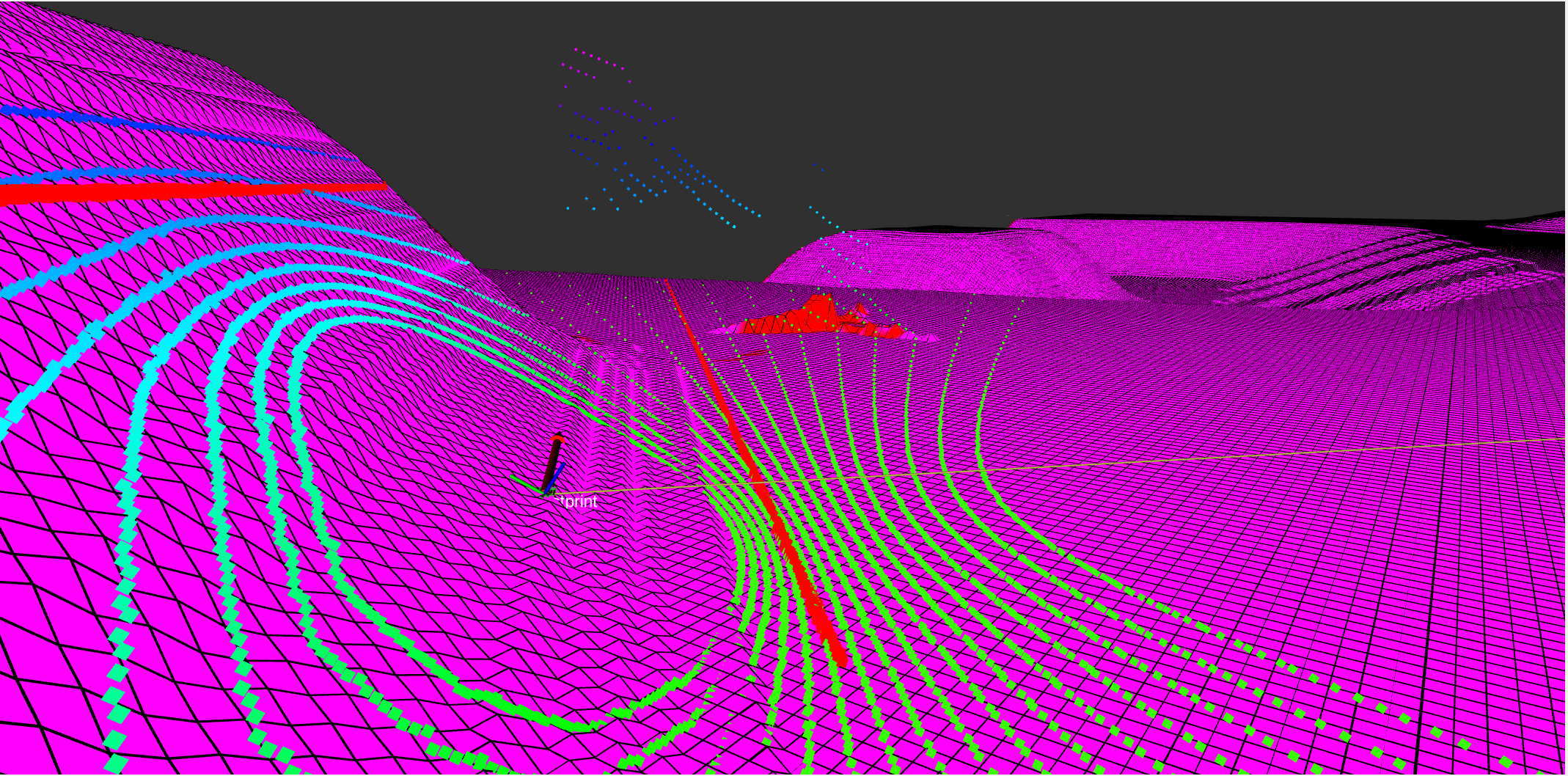}
  \caption{Robot in a non-planar environment equipped with LIDAR 2D and 3D using the popular visualization tool RViz. Octomap map (upper image) codifies 3D occupancy (colors indicate elevation in Z axe). Gridmap maps (bottom image) codify elevation and occupancy (colors indicate free space with fuchsia and obstacles with red. The green points are the obstacles detected by the 3D LIDAR, and the red points are those detected by the 2D LIDAR.}
  \label{fig:intro_uneven}
\end{figure}

Numerous challenges emerge when attempting to localize robots operating on uneven terrain, where the robot can be even inclined. Firstly, the robot's elevation (represented by the Z component of its position) must align precisely with the terrain elevation (Fig. \ref{fig:intro_uneven}). At the same time, its orientation in each axis is influenced by the terrain's inclination combined with data from an Inertial Measurement Unit (IMU), if available. Secondly, the representation of a three-dimensional world must encompass spatial occupancy, ensuring that regardless of the robot's orientation, we can accurately determine the expected distance perceived to obstacles. This is crucial whether the sensor beam detects the ground or a specific point on the surface of a complex object.

Presently, Nav2 employs 2D cost maps~\cite{6942636} referred to as Occupancy Grids (upper image in Figure \ref{fig:intro_uneven}. These grids represent the environment, organized into cells where each cell is assigned an occupancy value within the range [0-254], reserving the value 255 to denote unknown areas. Over the past two decades, this representation has effectively facilitated the navigation of thousands of ROS-compliant robots, operating as either Move Base~\cite{5509725} (ROS) or Nav2 (ROS 2). This representation is utilized for route planning, accommodating customizable safety margins, robot localization, and modeling dynamic and static obstacles. However, the inherent limitation of this representation is its assumption of flat terrain.

\begin{figure}[tb]
  \centering
  \includegraphics[width=0.91\linewidth]{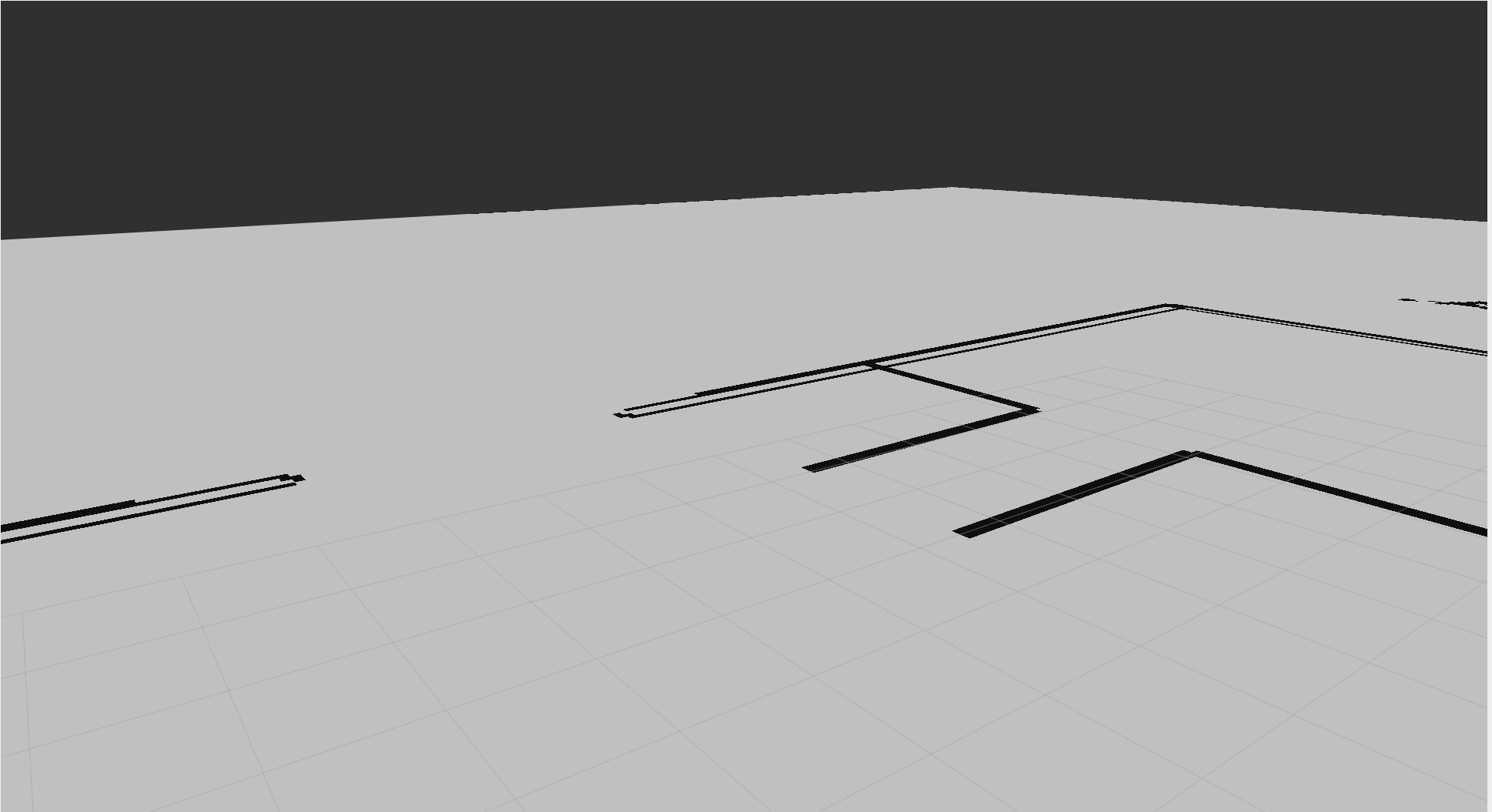}
  \includegraphics[width=0.91\linewidth]{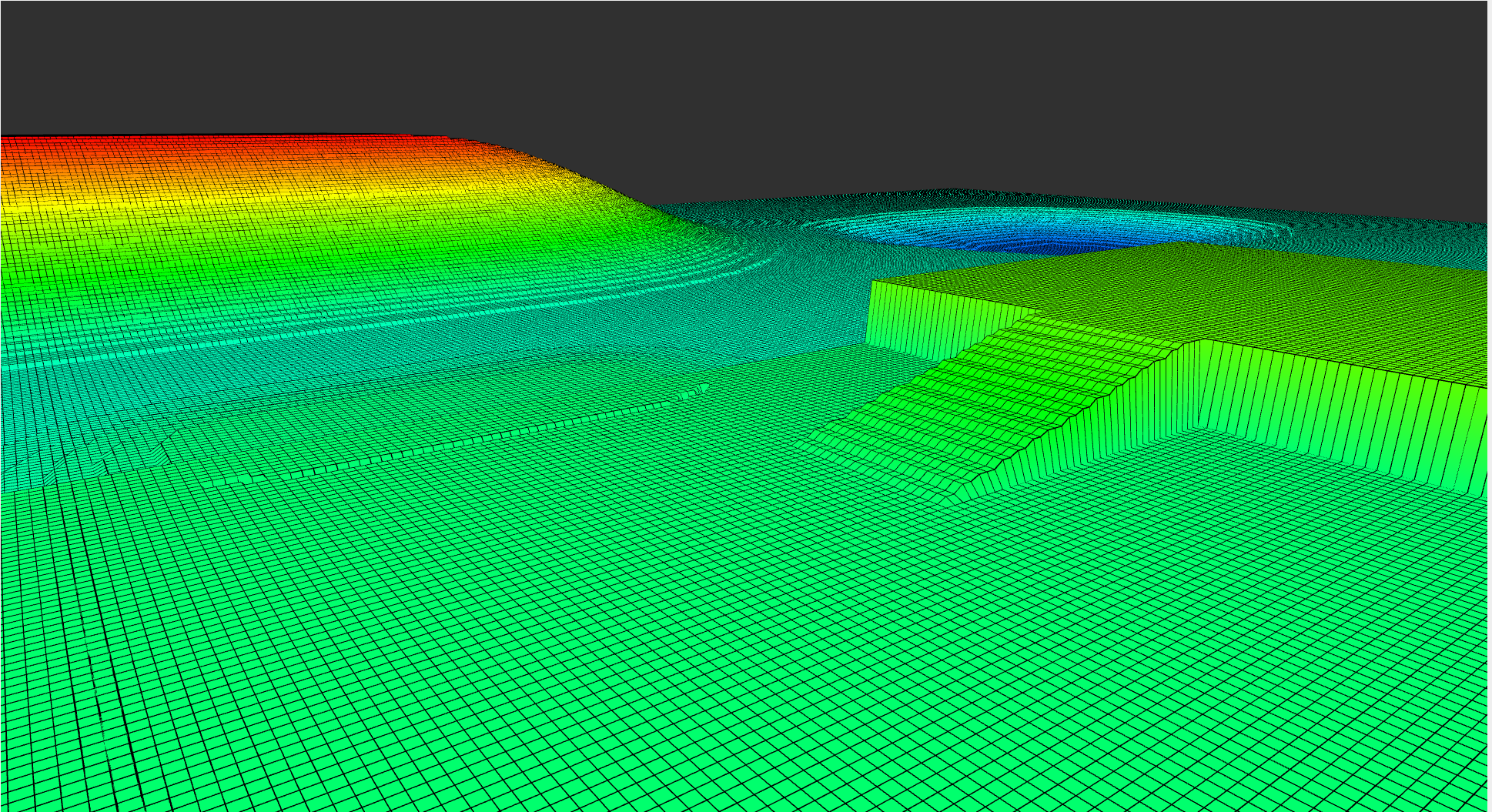}
  \caption{2D Occupancy gridmap (upper image, with free space in gray and obstacles in black) and the corresponding Gridmap map (bottom image, where colors indicate elevation in Z axe).}
  \label{fig:octogrid}
\end{figure}

In response to the challenges posed by non-planar environments, we advocate for the utilization of a composite map comprising two essential components: an Octomap~\cite{10.1007/s10514-012-9321-0}, which delineates the spatial occupancy of obstacles (as depicted in the upper image of Figure \ref{fig:octogrid}), and a Gridmap~\cite{Fankhauser2016GridMapLibrary}, which encodes crucial elevation and inclination data (illustrated in the lower image within the same Figure). Octomaps employ octrees\footnote{https://en.wikipedia.org/wiki/Octree} for the probabilistic representation of spatial occupancy. In contrast, gridmaps offer a versatile framework with multiple layers for encoding diverse information, including elevation, occupancy, or other relevant data within a 2D grid. Recognizing the significance of surface representation in robotic applications, we refrain from confining ourselves to use octomaps solely. Instead, we harness the capabilities of gridmaps, which allow us a continuous and gap-free representation, enabling the extrapolation of elevation data and the computation of inclinations.

Adaptative Monte Carlo Localization (AMCL) has emerged as the preeminent localization technique, renowned for its exceptional flexibility and reliability. Although other alternatives, such as Extended Kalman Filters (EKF)~\cite{9096805}, exist, AMCL has distinguished itself for its superior performance and widespread adoption, particularly within the Nav2 framework. Our approach uses gridmaps to estimate particles' position and inclination. Simultaneously, we utilize the octoMap to determine the likelihood associated with each individual particle. The calculation of this probability is contingent upon the readings from distance sensors, which emit 3D collision rays projecting from the sensor to the voxel corresponding to the space that should be occupied.

The motivation for this work is twofold, encompassing both scientific and technical aspects. On the technical front, we aim to contribute to the Open Source navigation community by offering a comprehensive, dependable, and efficient reference implementation. This implementation addresses the challenges of outdoor environments or scenarios featuring ramps and slopes within the Nav2 Framework. Moreover, we ensure backward compatibility to facilitate the integration of prior developments. Simultaneously, we pave the way for the community to explore opportunities for incorporating path planners, controllers, or alternative localization methods in situations where a 2D representation falls short of meeting the requirements.

This paper does not focus on the map creation process but rather assumes the availability of an existing map of the environment. In practical scenarios, numerous SLAM (Simultaneous Localization and Mapping) algorithms \cite{https://doi.org/10.1002/rob.21831}\cite{ye2019tightly} are readily accessible, capable of generating point clouds that can be conveniently processed to construct the combined map, comprising both the gridmap (containing elevation and occupancy information) and the octomap.

In summary, this paper presents several key contributions:
\begin{enumerate}
    \item The development of map representations for non-planar 3D environments incorporating elevation (gridmaps) and occupancy (octomaps) alongside their respective mapping processes.
    \item The paper introduces a localization technique grounded in Monte Carlo principles, distinguishing itself through its adaptability to non-planar environments featuring ramps or uneven terrain. This method bridges the divide between conventional 2D localization methods relying on preexisting maps and 3D approaches utilizing SLAM techniques but lacking such maps, thereby addressing a critical gap in the field. 
    \item Introduction of a novel observation model facilitating correction steps using both 2D and 3D sensors, applied to particles with 3D position/orientation and occupancy maps depicting obstacles.
    \item Implementation of these methodologies within Nav2, the reference navigation framework in ROS 2, ensuring comprehensive functionality and seamless integration. This standardization contributes to the field with methods that can be used for comparison in future research.
\end{enumerate}

We validate our proposed framework through experiments conducted in diverse settings, including simulation, outdoor environments, and indoor environments employing professional-grade robots.

The structure of this paper is organized as follows: Section \ref{sec:related} will provide an overview of related research in the field. Section \ref{sec:mcl} will offer a detailed exposition of our approach tailored for non-planar environments, representing the primary contribution of this paper. Section \ref{sec:nav} will introduce the architectural context of our framework. Experimental validation of our contributions and the results will be discussed in Section \ref{sec:exps}. We will provide some conclusions in Section \ref{sec:conclusions}.

\section{RELATED WORK}\label{sec:related}

The primary literature in this domain centers on Adaptive Monte Carlo Localization, commonly referred to as KLD-Sampling~\cite{Fox-1999-14948}, which is the current approach used within Nav2 and focused on localization in planar environments. It represents an evolution of the original Monte Carlo Algorithm~\cite{NIPS2001_c5b2cebf}. Since its maturation in the early 2000s, KLD-Sampling has emerged as the predominant localization algorithm, primarily owing to its remarkable robustness and flexibility~\cite{844766}, supplanting more conventional methodologies grounded in POMDPs~\cite{10.5555/1643031.1643040}, Markov~\cite{10.1007/10705474_1}, or occupancy grids~\cite{SCHIELE1994163}. This algorithm functions by sampling a probabilistic density function that characterizes the robot's position. Each sample, or particle, undergoes position updates provided by the perception model and weight adjustments governed by the observation model. Periodically, depending on the specific implementation, particles with lower weights may be replaced by others that closely align with those exhibiting higher weights. The capacity to disperse or reset particles across the environment makes AMCL a global localization method, affording the capability of initiation or reinitialization from inaccurate estimates. The computational overhead of the algorithm can be managed by regulating the number of particles~\cite{10160957}.

It is worth noting the significance of methods~\cite{maybeck1990kalman} rooted in Kalman Filters ~\cite{kalman1960}, which represent an optimal recursive estimator commonly employed in sensor fusion for system state estimation. In scenarios involving non-linear systems, which are prevalent in robotics, the Extended Kalman Filter (EKF) emerges as a preferred choice. EKF is inherently a local (or tracking) method, as it may face challenges when converging scenarios of complete uncertainty or erroneous position data. Nevertheless, some approaches address this limitation by combining multiple EKFs~\cite{MARTIN2007870}.

The research domain of localization within outdoor environments with dimensions exceeding those typically encountered in planar navigation is a relatively recent study area. Previous works, exemplified by \cite{4058483}\cite{5669675}\cite{4650585}, employ EKF-based methods to determine a robot's position in such environments by integrating data from GPS and other sensors. However, these approaches tend to overlook the critical aspect of environmental representation. They often operate under the assumption of relatively low precision and execute coarse-grained navigation strategies, which are ill-suited for precision navigation tasks. Subsequent evolution in this line of research entails leveraging a voxel representation combined with a particle filter to enhance the precision of autonomous vehicle navigation. Notably, our approach eliminates the dependency on GPS integration, rendering it capable of functioning independently of such devices. Alternative approaches\cite{1250626}\cite{1641929} have also been explored, focusing on point cloud mapping and neglecting GPS, relying solely on distance sensors. Recent works in this line, like \cite{9304679}, are motivated by the emergence of autonomous vehicles. They use a 3D representation of the environment based on voxel maps, similar to ours, but only for the perception model, discarding elevation maps or any other non-planar representation of the environment that can be used for the whole localization process.

An alternative approach to the localization problem has emerged from the pioneering work of \cite{4160954} focusing on 6D localization (position and orientation) of cameras. Although termed SLAM, these methods diverge from the traditional term SLAM in Mobile Robotics, as they do not produce maps for subsequent integration into a navigation system capable of exploration. Instead, they construct a set of 3D features solely to estimate camera motion between successive frames. This groundbreaking approach has progressed alongside SLAM algorithms in Mobile Robotics over the past two decades, with \cite{10160950} representing one of the latest advancements incorporating sophisticated features to anchor camera movements. However, the challenge with these methods in Robotics lies in their output, typically presenting as unstructured point clouds or, as seen in works such as \cite{https://doi.org/10.1002/rob.20209}, octomaps (kd-trees) requiring an EKF-based registration process. In our study, we use one of the latest approaches in this line \cite{lidarslam_ros2} to generate the initial point cloud representation of the environment (\ref{fig:campus_exp}). Unfortunately, these methods excel in achieving precise 6D odometry but fail in long-term robot localization within extensive environments due to eventual loop closure failures leading to map degradation.

Recent developments within the ROS framework have begun to tackle the challenge of robot navigation in non-planar environments. For instance, \cite{Fankhauser2014RobotCentricElevationMapping} introduced a solution tailored to quadruped robots navigating non-planar outdoor terrains. However, their focus primarily revolves around local mapping, incorporating elevation and obstacle information in the robot's immediate surroundings for obstacle avoidance. Unlike our proposal, their approach does not address global mapping or point-to-point navigation.
Similarly, in another recent work \cite{9981647}, the localization issue is not explicitly addressed, with GPS being presumed to provide location information. This approach involves planning robot routes directly on a map representing the environment as a raw point cloud. Notably, these non-flat terrain navigation solutions primarily assume outdoor terrains and rely on GPS for localization, even in scenarios with potential GPS signal loss or reduced precision.

Consequently, we believe that our approach represents a novel contribution to the state of the art by addressing non-planar navigation in GPS-denied or challenging indoor environments. It extends the scope beyond outdoor terrains and GPS-dependent solutions. While some recent work~\cite{https://doi.org/10.1049/iet-csr.2019.0043} explores navigation in underground environments where GPS signals are unavailable, it does not delve into the application of techniques like EKF or MCL, leaving potential avenues for future exploration. 


\section{MAP REPRESENTATION}\label{sec:map}

As previously noted, our research does not delve into the Simultaneous Localization and Mapping (SLAM) problem; instead, we presuppose the availability of a point cloud derived from the environment. This approach is deemed acceptable, given the abundance of prior works equipped with open implementations \cite{https://doi.org/10.1002/rob.21831}\cite{lidarslam_ros2} that can generate such point clouds. We firmly believe that incorporating existing works into our scientific endeavor, particularly when efficient implementations already exist, not only adds value but also underscores the notable progress made over the past decade in terms of interoperability and software reuse within the field of robotics.

Our starting point is a point cloud $\mathcal{PC}=\{p_0, p_1, \ldots, p_n\}$, with each point $p_i \in \mathcal{PC}$ having coordinates $(x, y, z)$ denoting occupancy. In indoor environments, it is imperative to incorporate all the information detected by the robot's distance sensors, regardless its orientation during operation. Typically, this encompasses the surrounding obstacles and extends to the ceiling and the floor within the point cloud data. 

As we introduced, our map comprises two integral components: an octomap and a gridmap, with the desired resolution $r$ (typically 0.1m is enough in most cases) obtained from the $\mathcal{PC}$:

\begin{enumerate}
    \item The generation of an octomap $\mathcal{OC}$ from $\mathcal{PC}$ is a straightforward process. In this procedure, each coordinate of a point within $\mathcal{PC}$ is designated as an occupied voxel in $\mathcal{OC}$. It is worth noting that this process can be expedited by reducing the density of $\mathcal{PC}$ using a certain resolution $r$.

%

    \item The generation of the gridmap $\mathcal{GR}$ uses $\mathcal{OC}$, beginning with an initial ground-level position defined at the onset of the procedure. Subsequently, a flooding algorithm is employed, wherein neighboring cells with a vertical (z-coordinate) difference below a specified threshold $thr$ are incorporated into the gridmap's elevation layer. Essentially, this layer delineates the robot's traversable positions within the environment. Additionally, an occupancy layer is added onto those gridmap cells that contain obstacles extending above the ground up to a designated height (typically corresponding to the robot's height), at a minimum. It's important to note that obstacles, such as walls or objects, are not represented in the gridmap through changes in elevation but rather as occupied regions within the occupancy layer. The elevation of these occupied cells remains consistent with neighboring non-occupied cells.






\end{enumerate}

\section{MONTE CARLO LOCALIZATION IN NON-PLANAR MAPS}\label{sec:mcl}

In the context of localization, Adaptive Monte Carlo Localization (AMCL) computes the probability distribution $bel(x_t)$, representing the robot's position at time $t$. This distribution is expressed as a collection $\mathcal{S}_t$ of hypotheses about the robot's pose $\mathcal{X}_t$ (as denoted in Equation ~\ref{eq:mcl1}).

\begin{equation}
\label{eq:mcl1}
\mathcal{X}_t = \{x^1_t, x^2_t, \cdots, x^I_t\}
\end{equation}

Every hypothesis, denoted as $x^i_t$ (where $1\leq i \leq I$), represents a possible position of the robot's state at time $t$. Each of these hypotheses within $X_t$ is associated with a corresponding weight, denoted as $w^i_t \in \mathbb{R}$, which quantifies its probability relative to all prior perceptions $z_{1:t}$ and actuation commands $u_{1:t}$ (as expressed in Equation \ref{eq:mcl2}).

\begin{equation}
\label{eq:mcl2}
w^i_t \sim P(x^i_t | z_{1:t}, u_{1:t}) 
\end{equation}

$\mathcal{P}_t$ represents the collection of \emph{particles}, where each \emph{particle} $p^i_t$ in $\mathcal{P}_t$ is a tuple ${<x^i_t, w^i_t, h^i_t>}$. In the subsequent section, we will delve into the details of $h^i_t$; but let set it aside for now. For simplicity, when referencing a \emph{particle} $p^i_t$, we can use the notation ${<x, w, h>}$.

A set of particles characterizes the robot's pose, providing an estimation that follows a normal distribution $\mathcal{N}(\boldsymbol{\mu}, \boldsymbol{\Sigma})$. Here, $\boldsymbol{\mu}$ denotes the mean and $\boldsymbol{\Sigma}$ the covariance matrix associated to $\mathcal{X}_t$.


\begin{figure}[tb]
  \centering
  \includegraphics[width=\linewidth]{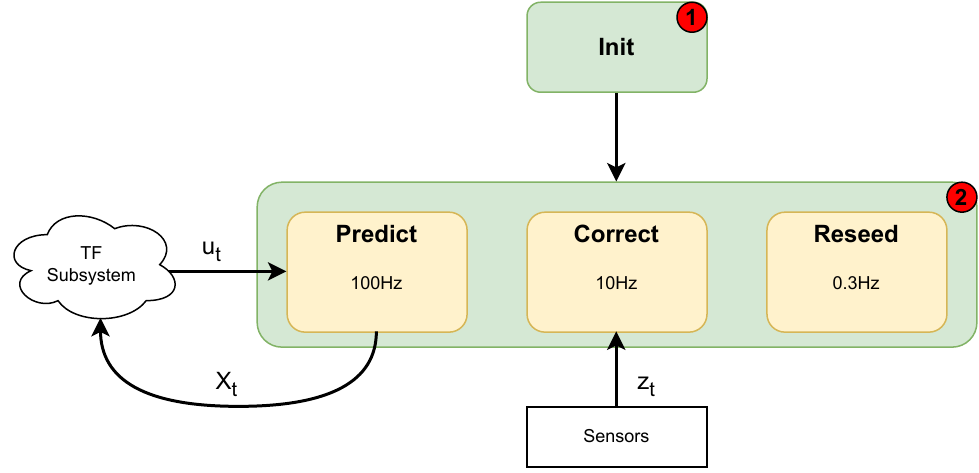}
  \caption{Diagrams of the distinct processes that update $\mathcal{P}_t$. After the initialization, three synchronous processes update the particles at different rates.}
  \label{fig:flowchart}
\end{figure}

The AMCL algorithm is structured into three distinct processes, as shown in Figure \ref{fig:flowchart}, each aimed at updating the particles to incorporate the information from $z_t$ and $u_t$: 
\begin{itemize}
\item \textbf{Prediction}: This process focuses on updating the positions of the particles $\mathcal{X}_t$ in response to the observed displacement $u_t$.
\item \textbf{Correction}: In this process, the weights of the particles $\mathcal{W}_t$ are adjusted based on the sensor readings $z_t$.
\item \textbf{Reseed}:  In this process, hypotheses $x^i_t$ with weights $w^i_t < threshold$ are eliminated, and new hypotheses are generated near those with higher weights.
\end{itemize}

In the original algorithm, these three phases are conventionally executed continuously in sequence. However, in our modified approach, we perform each phase independently at varying frequencies to optimize computational efficiency while minimizing the robot's computational load. To provide a frame of reference, we execute the prediction phase at 100Hz, the correction phase at 10Hz, and the reseed phase at 0.3Hz. Position updates are performed at a higher frequency to ensure optimal accuracy when making navigation decisions. In contrast, particle weight updates, representing the most computationally intensive phase, occur less frequently. Conducting the reseed phase at a higher frequency is deemed unnecessary without compromising effectiveness.

Adaptive Monte-Carlo Localization dynamically adjusts the number of particles in response to the uncertainty associated with $\mathcal{X}_t$, whether the particles are concentrated or dispersed. When uncertainty is low, the number of particles is reduced and increases when uncertainty rises.



Our research extensively relies on a geometric transformation system known as TF~\cite{6556373}, which meticulously preserves the relationships (translation and rotation) between frames, also referred to as reference axes. This system enables the transformation of coordinates from one frame to another, provided they are interconnected within the same TF tree. The TF tree structure links frames in such a way that each of them possesses a sole parent frame while being capable of accommodating multiple child frames.

\begin{figure}[tb]
  \centering
  \includegraphics[width=0.49\linewidth]{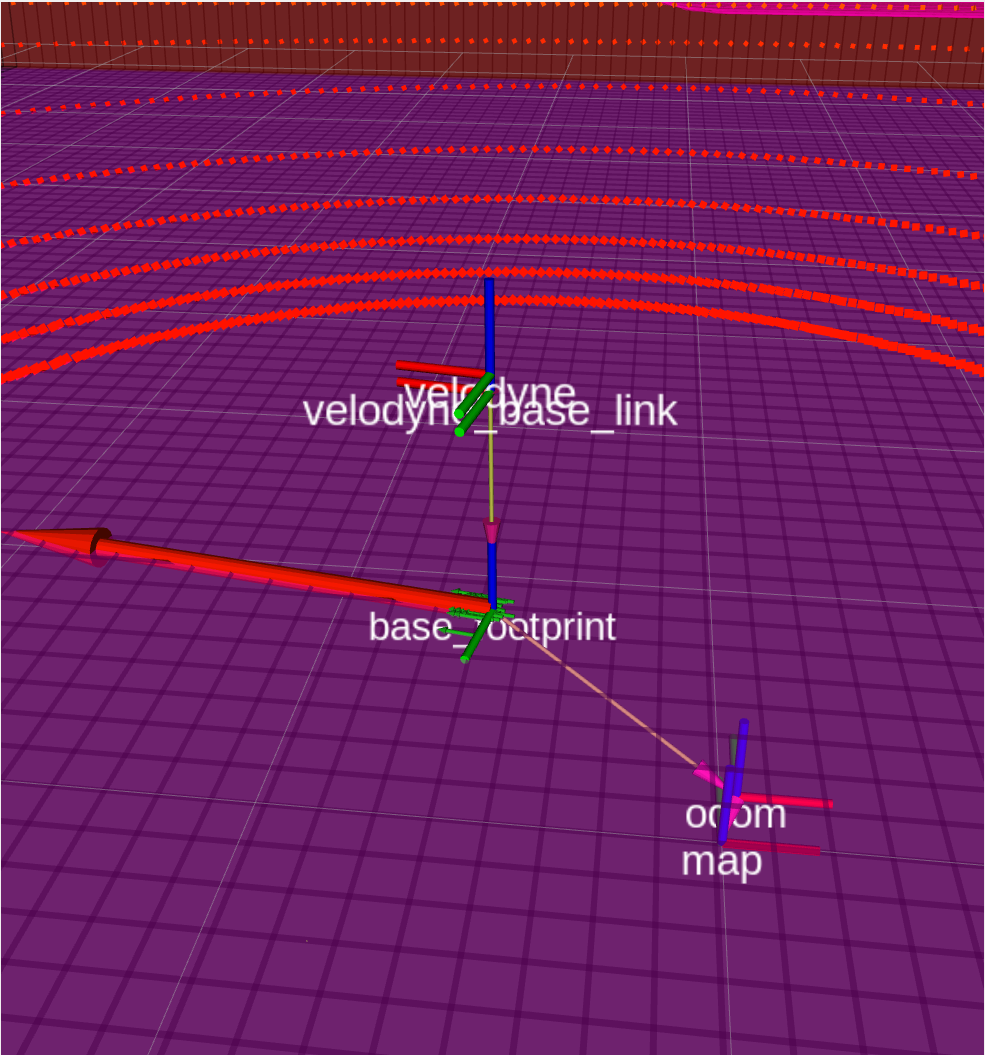}
  \includegraphics[width=0.49\linewidth]{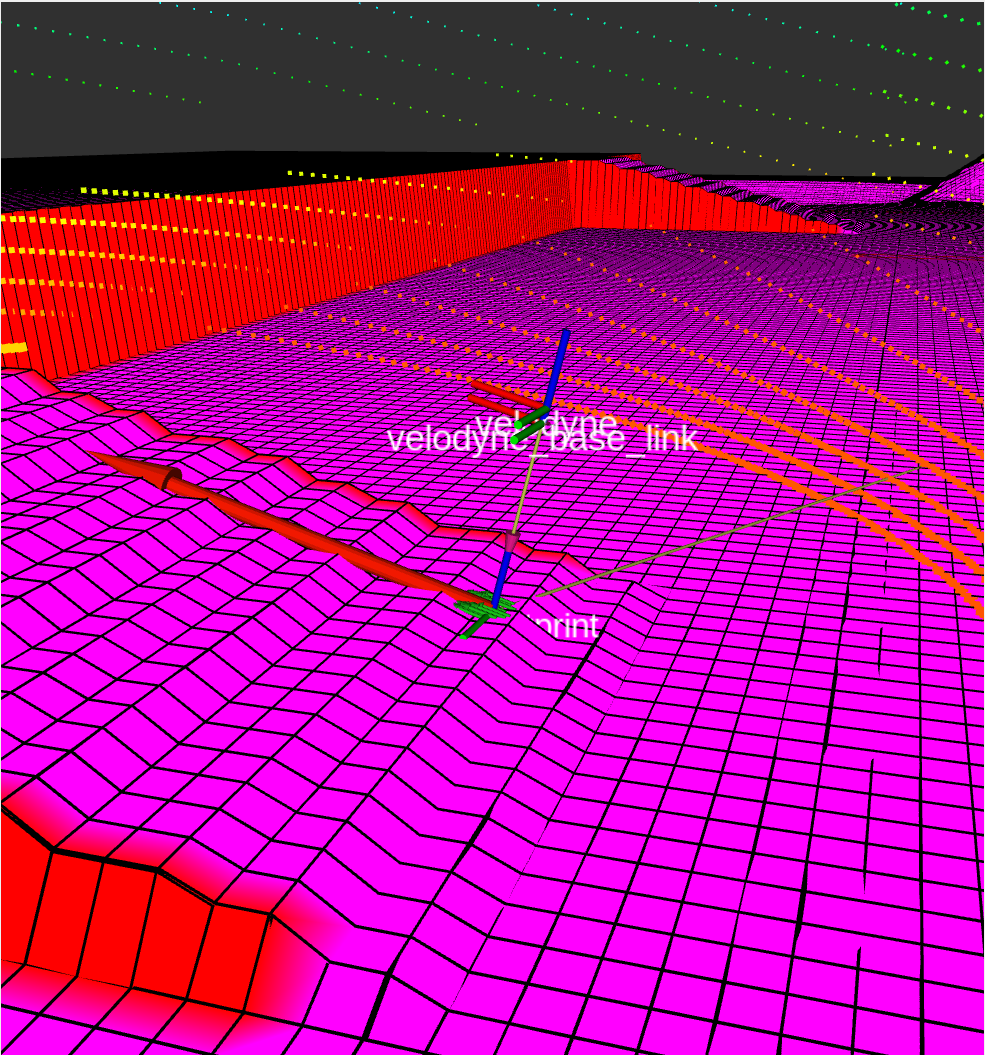}
  \caption{Some frames of the robot's TF tree. The robot's inclination is provided by the transform generated by the robot's driver.}
  \label{fig:tf_mov}
\end{figure}

As illustrated in Figure \ref{fig:tf_mov}, the frame representing the robot is denoted as \emph{base\_footprint} (abbreviated as \emph{bf}). A connection exists from \emph{bf} to the frame housing all laser readings, designated as \emph{base\_scan}. The \emph{odom} frame indicates the robot's initial position. The \emph{odom$\to$bf} transform encodes the robot's displacement since its initialization. The \emph{map} frame serves as the parent frame for \emph{odom}. While a localization algorithm computes the $map \to bf$ relationship, the subsequent relationship established, after subtracting $odom \to bf$, becomes $map \to odom$.

The TF system receives rotation and translation relationships among frames, yielding a $6 \times 6$ transformation matrix, denoted as $RT^t_{A \to B}$, to facilitate the transition between frames from distinct sources, some operating at exceedingly high speeds. For brevity, we will employ an alternative notation, namely $A2B_t$. This system accommodates requests for relationships denoted as $X2Y_{t'}$ at a specified time $t'$, even if frames $X$ and $Y$ lack a direct connection or if precise information is unavailable at time $t'$. The TF system employs interpolation to provide the necessary transformations in such cases.

\subsection{Prediction}
\label{sec:prediction}

The goal of this phase is to update the position of each particle $x^i_t\in X_t$
with the detected displacement $u_t$. If the last prediction was made in $t-1$, the displacement $u_t$ is the difference
of the relations $odom2bf$ from time $t-1$ and current time $t$ (Equation \ref{eq:mcl3}).

\begin{equation}
\label{eq:mcl3}
    u_t = odom2bf_{t-1}^{-1} * odom2bf_t
\end{equation}

Each particle is updated (Equation \ref{eq:mcl4}) using $u_t$ plus a random noise $e(u_t)$ that follows a normal distribution $\mathcal{N}(0, E_u)$, being $E_u$ a parameter known \emph{a priori} that represent the odometry accuracy.

\begin{equation}
\label{eq:mcl4}
    x^i_t = x^i_{t-1} * (u_t + e(u_t))
\end{equation}


%
%

Another important correction, which occurs as the robot traverses the environment, involves adjusting the z-coordinate of each particle to match the elevation data stored in the gridmap at the particle's position. While we could have employed this inclination information to correct the robot's orientation, we used the inclination information derived from the TF transformation $map \to bf$ (Figure \ref{fig:tf_mov} instead. This choice was made since the TF system integrates odometry and IMU information. Nevertheless, it's worth noting that using the inclination data from the gridmap remains a viable option, especially in cases where the robot lacks access to such sensors.

\subsection{Correction}
\label{sec:correction}

During this phase, the observation model updates $W_t$ for each particle $x^i_t\in X_t$. Beginning with a comprehensive set of sensor readings denoted as $Z_t$, where each reading is represented as $z^j_t \in Z_t$, we update $w^i_t$ based on the calculation of $p(z^j_t|x^i_t)$. In simpler terms, for every particle and each laser reading, we compare the obtained reading from the sensor with what would have been expected at the particle's hypothetical position, using the octomap $\mathcal{OC}$. If they closely align, the weight of the particle increases. Conversely, a significant disparity would suggest that the robot should not have detected this reading at that particular hypothetical position, reducing the particle's weight.

\begin{figure*}[bh!]
  \centering
  \includegraphics[width=1\linewidth]{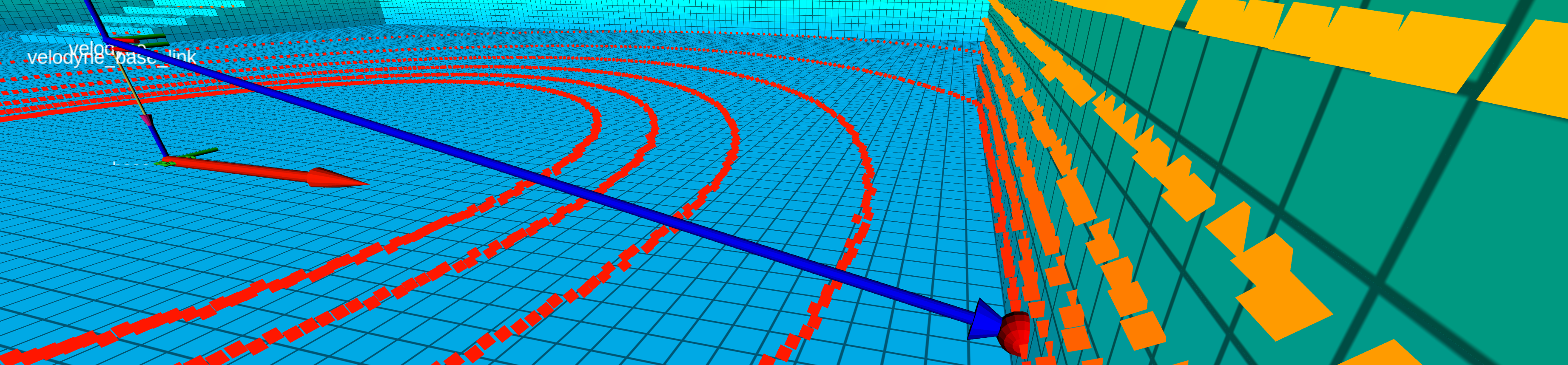}
  \caption{Ray that links the origin of a 3D laser sensor reference axes with the position of one of its readings. The red sphere indicates the collision of the beam with the octomap.}
  \label{fig:percept}
\end{figure*}

In this study, we have integrated multiple distance sensors, including 2D lasers, 3D lasers, and RGBD cameras. Every $z^j_t \in Z_t$ is a coordinate $(x, y, z)$ indicating a detection of an obstacle in the sensor frame. 
Each sensor comes with its own specifications pertaining to the precision of observations, with some sensors exhibiting higher precision than others. For sensors that generate a substantial amount of data, such as 3D lasers or RGBD cameras, we have the flexibility to determine the number of readings to be discarded for each one utilized. This approach effectively alleviates the CPU load without compromising observation quality as long as the discard rate remains within reasonable bounds.

A common step to be applied across all sensors involves a preliminary process where all the readings are consolidated, treating them as 3D points to detect obstacles. This unification allows us to consistently reconstruct the lines connecting each sensor to the perceived obstacles, as the points from each sensor are inherently referenced to their respective coordinate axes.

In the correction step (Algorithm \ref{alg:correct}), each particle $p^i_{t} \in \mathcal{P}_t$ updates its probability $p^i_{t}.w$, comparing the sensory reading $Z_t$ with the one that should have been obtained if the robot was really in $\mathcal{P}_{t}.\mathcal{X}$. This comparison is made for each reading $z^j_t$ and for each particle $p^i_{t}$, using the Bayes theorem as shown by equation \ref{eq:mcl5} and \ref{eq:mcl6}.

\begin{equation}
\label{eq:mcl5}
    p^i_{t}.w = P(p^i_{t}.x | z^j_t) = \frac{P(z^j_t | p^i_{t}.x) * P(p^i_{t}.x)}{P(z^j_t)}
\end{equation}

\begin{equation}
\label{eq:mcl6}
    p^i_{t}.w = P(z^j_t | p^i_{t}.x) * p^i_{t}.w = \frac{1}{\sigma * \sqrt{2 * \pi}} e^{-\frac{1}{2}(\frac{error}{\sigma})^2} * p^i_{t}.w
\end{equation}

The $\sigma$ parameter is a value that represents the sensor precision, and it is known \emph{a priori}. The \emph{error} value is the difference between the measured distance and the theoretical distance (Equation \ref{eq:mcl7}).

\begin{equation}
\label{eq:mcl7}
error = |z^j_t.dist - z'^j.dist|
\end{equation}

\begin{algorithm}[h!]
\caption{Correction step}\label{alg:correct}
\begin{algorithmic}[1]

\Function{Correction}{$\mathcal{P}_{t}$, $Z_t$}
    \State $bf2sensor^t \gets RT^t_{bf \to sensor}$
    \ForAll{$z^j_t$ in $Z_t$}
        \ForAll{$p^i_{t}$ in $\mathcal{P}_t$}
            \State $z'^j.dist \gets theoretic\_distance(p^i_{t}, z^j_t, bf2sensor^t, \mathcal{OC})$
            \State $z^j.dist \gets \sqrt{{z^j_x}^2 + {z^j_y}^2 + {z^j_z}^2} $

            \State $p^i_{t}.possible\_hits \gets p^i_{t}.possible\_hits + 1$
            \If{$z'^j.dist == \infty$ and $z^j.dist == \infty$}
                \State $p^i_{t}.h \gets p^i_{t}.hits + 1$
            \Else
                \State $error = |z^j_t.dist - z'^j.dist|$
                \State $p^i_{t}.w = p^i_{t}.w * P(p^i_{t}.x | z^j_t) = $
                \State \hspace{1cm} $\frac{P(z^j_t | p^i_{t}.x) * P(p^i_{t}.x)}{P(z^j_t)} = $
                \State \hspace{1cm} $P(z^j_t | p^i_{t}.x) * p^i_{t}.w$ =
                \State \hspace{1cm} $\frac{1}{\sigma * \sqrt{2 * \pi}} e^{-\frac{1}{2}\frac{error}{\sigma}^2} * p^i_{t}.w$
                \If{$error < threshold$}
                    \State $p^i_{t}.h \gets p^i_{t}.hits + 1$
                \EndIf
            \EndIf
        \EndFor
    \EndFor
    \State normalize($\mathcal{P}_t.w$)
\EndFunction

\end{algorithmic}
\end{algorithm}

\begin{figure*}[bh!]
  \centering
  \includegraphics[width=1\linewidth]{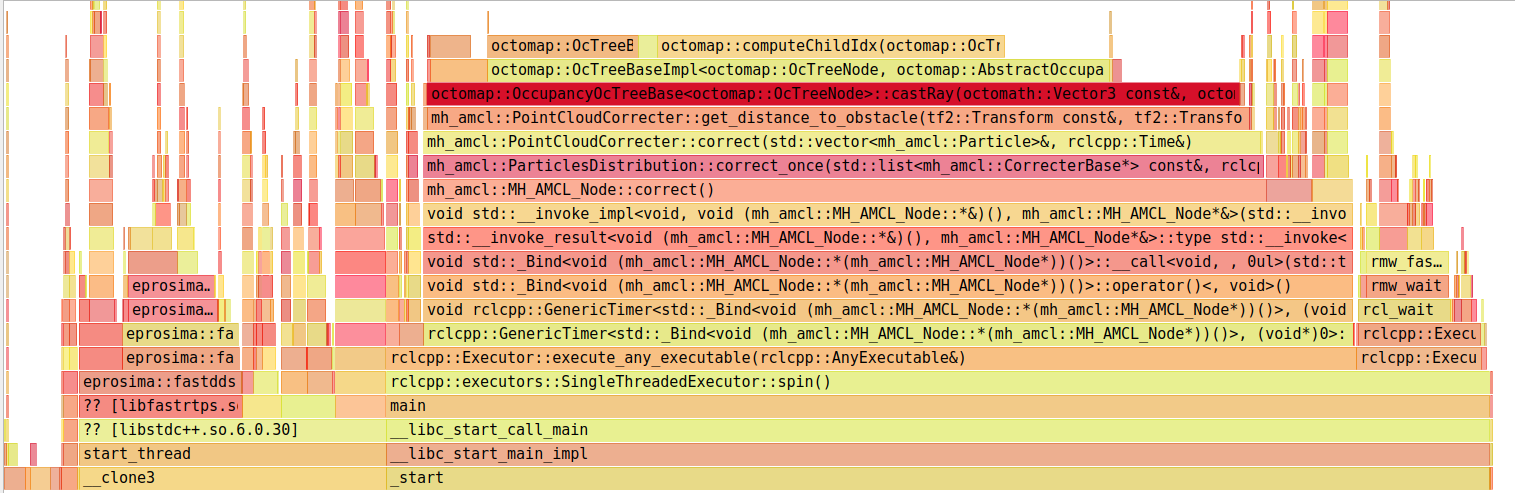}
  \caption{Profile of the execution of the algorithm, using the tool Hotspot\cite{hotspot}. The Octomap's \texttt{castRay} function takes most of the computation time for creating all the raytraces.}
  \label{fig:hotspot}
\end{figure*}

The value \emph{theoretic\_distance} calculated in Algorithm \ref{alg:correct} is a raytrace going from the sensor origin $p^i_{t} * bf2sensor^t$ to the sensor reading $p^i_{t} * bf2sensor^t * z^j_t$ within $\mathcal{OC}$. Notably, the computation time required for this operation during a correction step is significantly resource-intensive, as shown in Figure \ref{fig:hotspot}. In the specific octomap implementation we use, we mitigate this processing time by leveraging parallelization libraries such as OpenMPI~\cite{10.1007/11752578_29}. 

Each particle within $\mathcal{P}_t$ does not only is composed by a position denoted as $p^i_{t}.x$ and a weight represented as $p^i_{t}.w$, but also features a \emph{hits} field denoted as $p^i_{t}.hits$ and a \emph{possible\_hits} field marked as $p^i_{t}.possible\_hits$. These additional fields are used to calculate each particle's likelihood compared to the most recent perception, indicating how frequently the laser readings would encounter obstacles if the robot were positioned according to the particle's pose. The effectiveness of $\mathcal{P}_t$ is assessed using these fields, as expressed in Equation \ref{eq:mcl9}. In the validation section, we will demonstrate how this value provides a more descriptive and reliable measure than relying on the covariance matrix.

\begin{equation}
\label{eq:mcl9}
Quality(\mathcal{P}_t) = \frac{\sum_{i=0}^I \frac{p^i_{t}.hits}{p^i_{t}.possible\_hits}}{|\mathcal{P}_t|}
\end{equation}

\subsection{Reseed}
\label{sec:reseed}

\begin{figure}[tb]
  \centering
  \includegraphics[width=0.95\linewidth]{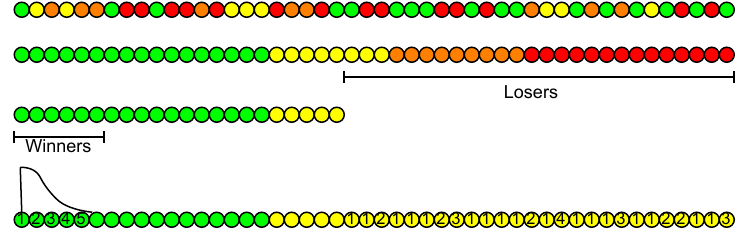}
  \caption{Reseed. Particles' color indicates high probability (green) to low probability (red).}
  \label{fig:reseed}
\end{figure}

In the reseeding process, particles with lower weights are substituted with particles close to those with higher weights; this procedure is governed by two key parameters: the \emph{percentage of winners} and the \emph{percentage of losers}. This process unfolds through the following steps:

\begin{enumerate}
    \item The population of particles $\mathcal{P}t$ is sorted based on weights $p^i{t}.w$, distinguishing between \emph{winners} and \emph{losers} based on given percentage thresholds. The remaining particles are commonly referred to as \emph{no-losers} (Figure \ref{fig:reseed}).
    \item \emph{Loser} are removed from the population.
    \item Those particles that have been removed are inserted back, each one randomly selecting a replacement from the pool of \emph{winners}, following a normal distribution $\mathcal{N}(0, winners/2)$.
\end{enumerate}

Additionally, this step allows us to control the number of particles within the range of $[particles_min, particles_max]$. If the distribution's covariance is high, we increase the number of particles; conversely, if it is low, we decrease it. This adaptive adjustment mitigates computational load when the distribution converges to a particular position.

\section{OPEN-SOURCE NAVIGATION INTEGRATION}\label{sec:nav}

Although successfully implementing an algorithm or localization system for a specific robot can advance knowledge within a particular domain, it often remains confined to the laboratory or robot for which it was originally designed. True scientific progress demands research that can be replicated by fellow scientists who may have access to similar resources, if not identical ones. We firmly believe that efforts should be dedicated to translating research into software that peers can execute, and this is precisely where Open Source initiatives prove invaluable. Establishing common standards is essential for building a foundation for an entire scientific community to collaborate.

In recent years, the advent of ROS has introduced a widely embraced standard, enabling seamless integration of software developed by various contributors. ROS provides established procedures, practices, and standards familiar to the broader ROS community. A ROS developer possesses the know-how to build, configure, and execute ROS-compliant software. In the realm of navigation, ROS has played a pivotal role, initially with \emph{move\_base}~\cite{5509725} and now with \emph{Nav2}~\cite{macenski2020marathon2}, by offering a framework for the development and integration of localization and navigation algorithms. This framework has facilitated the navigation of thousands of robots with diverse locomotion methods, sensors, and morphologies.

For instance, developing a route planning algorithm within Nav2 ensures that it becomes accessible to the wider robot navigation scientific community, fostering its utilization and enabling comparisons with both past and future approaches to the same problem.

\begin{figure}[tb]
  \centering
  \includegraphics[width=0.95\linewidth]{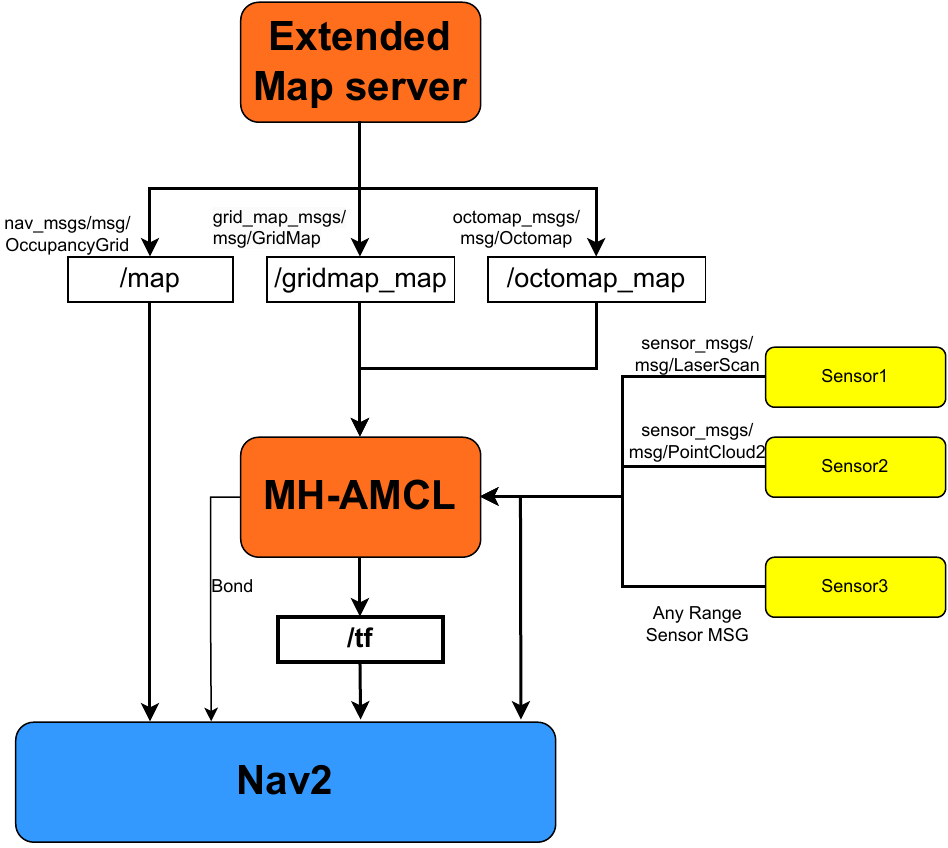}
  \caption{Simplified scheme of integration in Nav2. Our contributions are the orange boxes.}
  \label{fig:innav2}
\end{figure}

Hence, we have dedicated substantial efforts to incorporate the contributions detailed in this document into Nav2, rendering them accessible to the navigation scientific community as Open Source resources~\cite{mhamcl}\cite{mapserver}\cite{materialpaper}. Figure \ref{fig:innav2} illustrates the components we have developed and their interconnections with each other and with the Nav2 framework: MH-AMCL\footnote{The acronym \emph{MH} is retained for historical reasons, denoting its multi-hypothesis nature. This nomenclature stems from our work's evolution from prior research~\cite{10160957} that was designed for flat environments and supported multiple hypotheses. It's worth noting that this particular option is not supported in the current work presented in this document.} and Extended Map Server.

Our approach involves the replacement of existing modules in Nav2 with our own modules while preserving interfaces and adhering to Nav2 component requirements. For example, a bond connection between MH-AMCL and Nav2 serves as a mechanism, consistent with the standard practice across Nav2 modules, to signal their operational status and thereby detect potential system failures.

\begin{itemize}
    \item \textbf{Extended Map Server}: The original Map Server exclusively publishes a 2D occupancy grid to the \texttt{/map} topic. In contrast, our Extended Map Server extends it by generating the $\mathcal{OC}$ octomap and publishing it to \texttt{/octomap\_map}, as well as producing the $\mathcal{GR}$ Gridmap and making it available on \texttt{/gridmap\_map}. Furthermore, the server continues providing the essential 2D occupancy grid directly from the occupancy layer in $\mathcal{GR}$, published on \texttt{/map}. The inclusion of \texttt{/map} is crucial for Nav2, as all of its current navigation algorithms rely on the 2D map, necessitating the preservation of backward compatibility. As Nav2 evolves and introduces navigation algorithms that require $\mathcal{OC}$ or $\mathcal{GR}$ information, the Extended Map Server will provide the additional data required by Nav2.
    \item \textbf{MH-AMCL}: MH-AMCL replaces AMCL, the core self-localization component within Nav2. In Figure 1, certain interfaces required for a Nav2 localization method have been concealed, such as publishing the particles, the robot's position, its covariance, or subscribing to potential robot position resets. The critical interfaces that MH-AMCL engages with include subscriptions to various range sensors and the maps generated by the Extended Map Server. The primary output of the localization module is the generation of the TF $map \to odom$, which encapsulates the robot's positional information on the map (see REP-105~\cite{rep105} for frame standards).
\end{itemize}

This integration endeavor has yielded fully operational modules seamlessly incorporated into Nav2, offering usability to companies seeking to employ them in their projects and providing scientists with a benchmark for comparative research. In the subsequent section, we assess our approach in comparison to the original AMCL, a process facilitated by the adherence of both components to the Nav2 integration standards.

\section{EXPERIMENTAL VALIDATION}\label{sec:exps}

We conducted a series of experiments to validate our approach across diverse robotic platforms and environmental scenarios. The primary objective was to ascertain the reliability and accuracy of our method while also evaluating the hypothesis that it outperforms the reference implementation of AMCL in Nav2.

In all the experiments, we executed our approach in a computer with a 12th Gen Intel Core i7-1280P with 20 cores and 48 GB RAM.

The validation process employed the following metrics:

\begin{itemize}
    \item \textbf{Translation and Rotation Error}: We quantified the error in both translation and rotation, assessing the robot's 6D position and orientation relative to a Ground Truth system. This metric is pivotal in validating our algorithm's performance. Our initial target was to maintain errors below 30 centimeters in most instances. However, we acknowledge that slightly higher error thresholds may still be deemed acceptable in outdoor settings or situations with limited perceptual data.
    \item \textbf{Quality and Uncertainty}: As previously discussed, it is common practice to gauge the accuracy of a robot's position estimate by examining its level of uncertainty. Our approach incorporates an alternative quality metric derived from the average alignment of sensory readings with those expected at each particle's estimated position. We found that this metric offers improved insights, particularly in phases characterized by elevated error rates.
    \item \textbf{Computational Time}: A localization method's utility depends on its ability to deliver accuracy without excessively burdening the robot's computational resources. To this end, we measured the resource consumption of each phase within the algorithm and the overall computational time.
\end{itemize}
Our experimental validation is conducted in three distinct scenarios to assess the performance of our approach comprehensively:

\begin{enumerate}
    \item Simulated Indoor/Outdoor Environment: This scenario (Figure \ref{fig:setup_1}) encompasses various terrain features, including ramps, elevations, and depressions. For this evaluation, we employ a Summit XL robot outfitted with both a generic 2D lidar and a 3D lidar sensor.

\begin{figure}[h!]
  \centering
  \includegraphics[width=\linewidth]{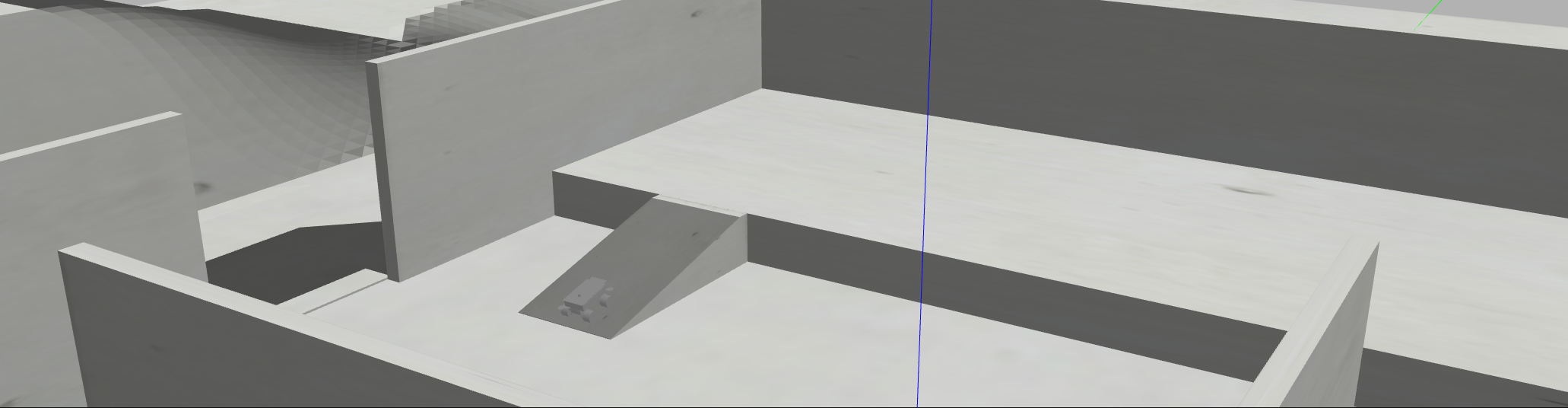}
  \includegraphics[width=\linewidth]{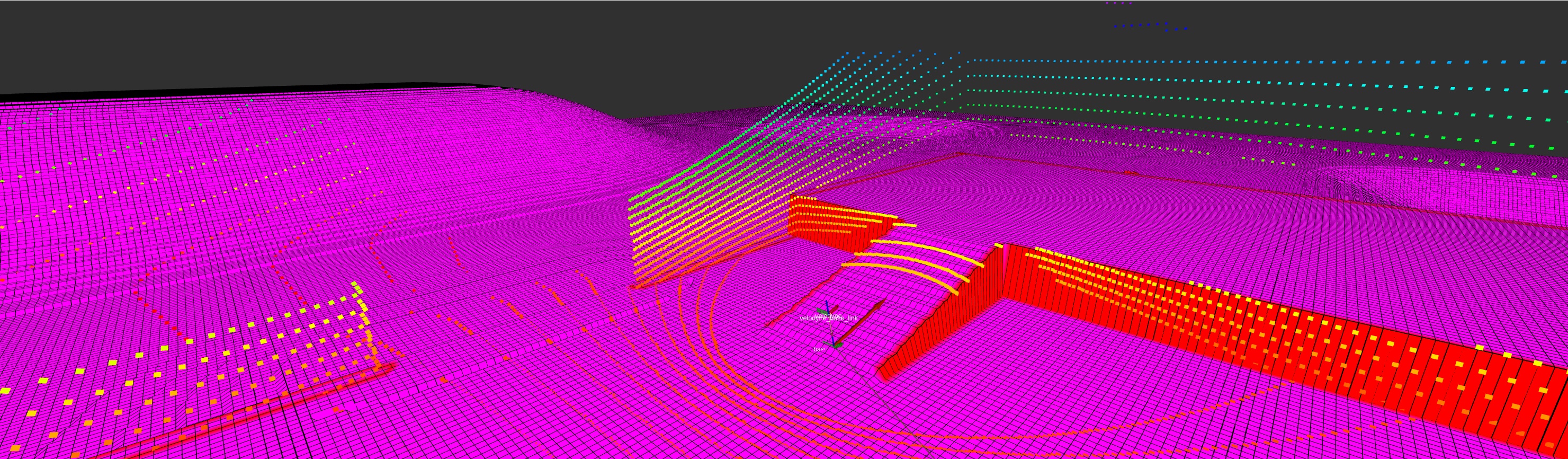}
\caption{Setup for the experiment in simulation. A Summit XL robot equipped with laser 2D and 3D in an indoor/outdoor environment with slopes. The upper image shows the simulation with the robot. The bottom image shows the elevation map, the robot pose, and the laser 3D perception.}
  \label{fig:setup_1}
\end{figure}

    \item Real Indoor Environment (Laboratory): The second scenario involves a real-world indoor setting, specifically our laboratory (Figure \ref{fig:setup_2}), where we deploy a Tiago Robot, a professional-grade robot with a differential base and a 270º 2D laser Hokuyo UST-10LX.

\begin{figure}[h!]
  \centering
  \includegraphics[width=\linewidth]{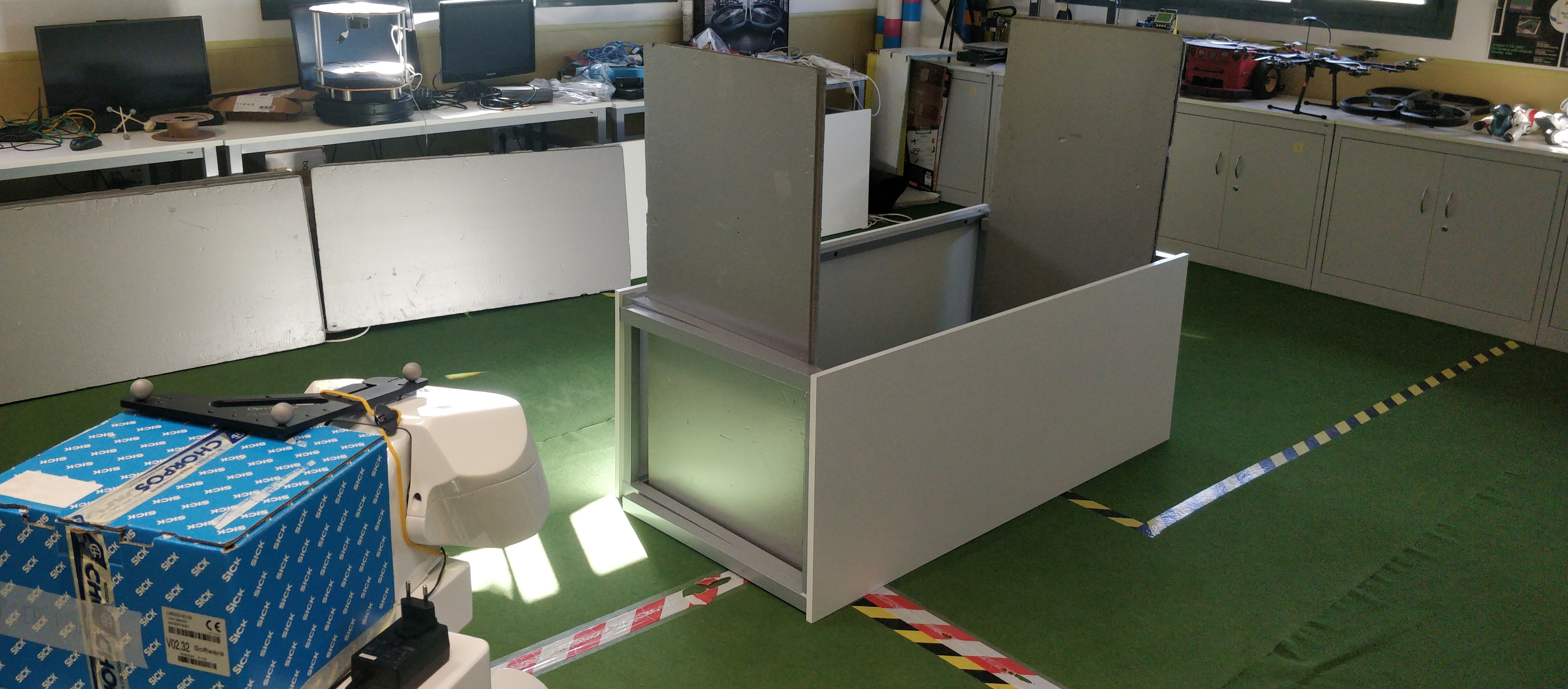}
  \includegraphics[width=\linewidth]{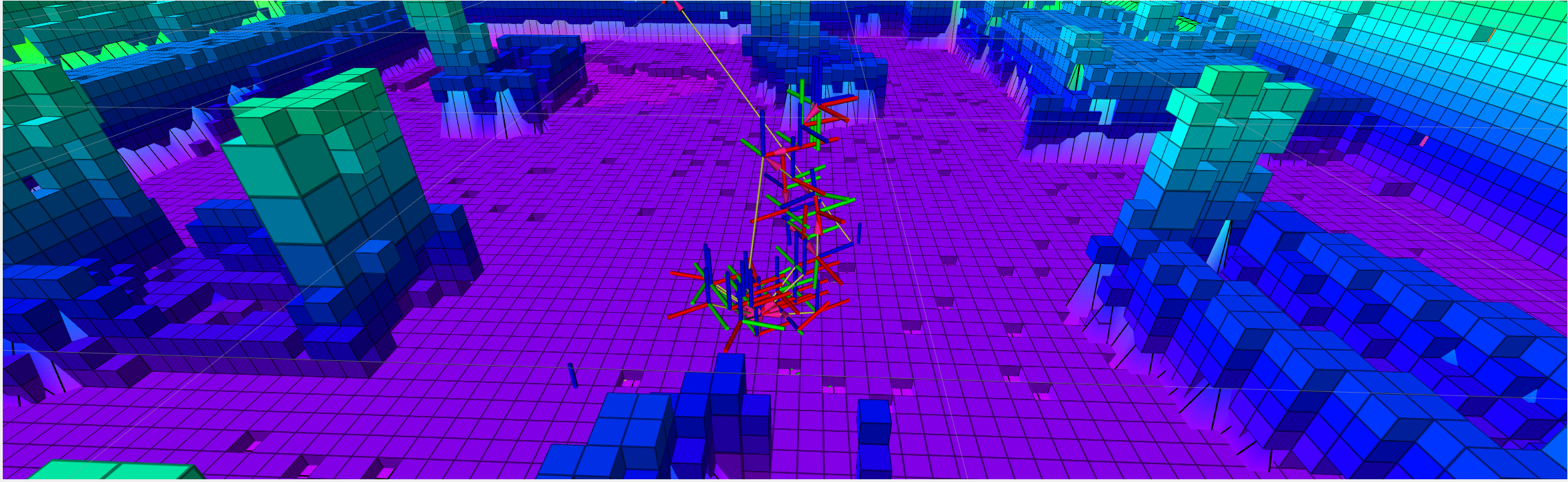}
\caption{Setup for the experiment in a real robot. A Tiago robot equipped with laser 2D. The upper image shows the robot with the mocap marker. The bottom image shows the TFs and octomap.}
  \label{fig:setup_2}
\end{figure}

    \item Real Indoor/Outdoor Environment (Campus): The third scenario involves a real-world scenario using the actual Summit XL robot equipped with a (upper image at Figure \ref{fig:robot_rampa}), performing a teleoperated navigation predominantly outside the laboratory building, with the concluding segment inside the facility. The primary sensor that is equipped with the robot is the 3D Laser Robosense RS HELIOS 5515, which creates a point cloud that captures obstacles up to a distance of 150 meters. This robot has 4 wheels but works as a differential robot: the robot turns applying opposite velocities to the wheels of each side, producing noisy odometry.
    The building’s exterior features irregular terrain along the campus walks, including ramps in certain areas (bottom image at Figure \ref{fig:robot_rampa}).

\begin{figure}[h!]
  \centering
  \includegraphics[width=\linewidth]{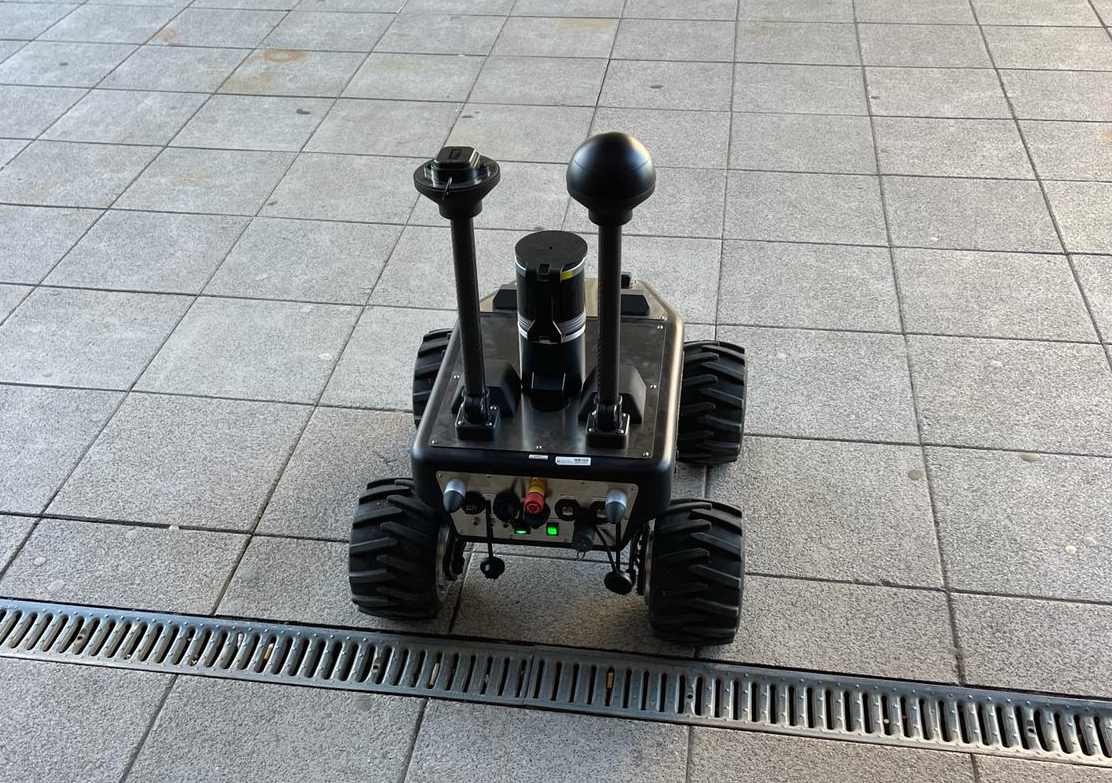}
  \includegraphics[width=\linewidth]{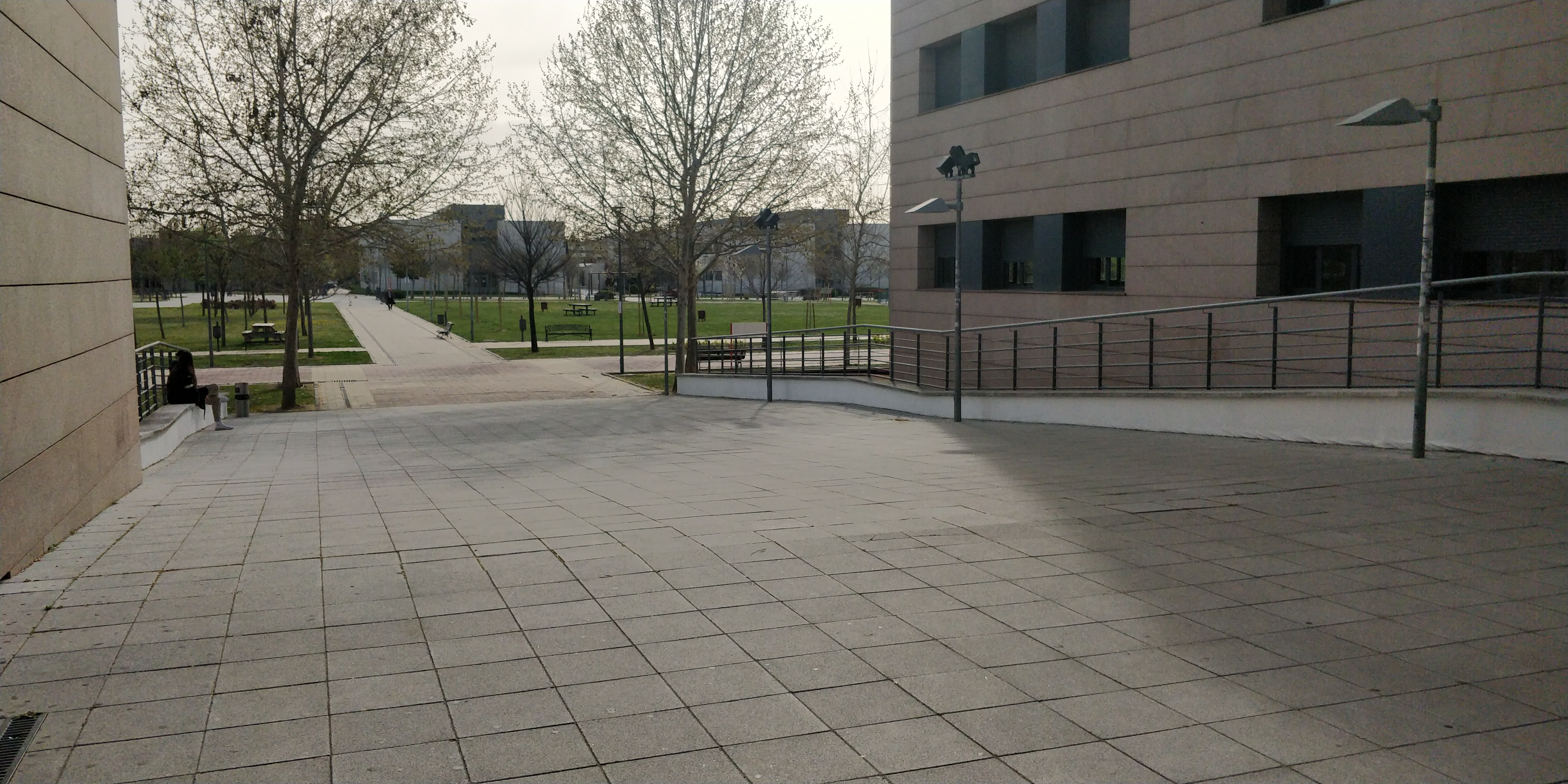}
\caption{Summit XL equipped with a 3D Laser (up) and one of the ramps in the environments (bottom).}
  \label{fig:robot_rampa}
\end{figure}

\end{enumerate}


For the first two experiments, we will leverage our motion capture framework, MOCAP4ROS2~\cite{10.1145/3555051.3555076}, which facilitates the detection of markers attached to the robot. These markers enable us to precisely determine their positions, including translation and orientation, using a camera system. Importantly, the markers's position relative to the robot's frames is accurately known. MOCAP4ROS2 seamlessly integrates with ROS 2, enabling the reception of ground truth information from the robot via topics that include timestamps.

\subsection{Experiments in simulation}

Our experimental validation comprises simulation experiments, offering the advantage of deploying various robot configurations and sensor setups within an idealized environment to test our localization system comprehensively. The robot's position in the simulation is determined using the MOCAP4ROS2 framework simulator plugin.

For conducting these experiments, we employ the rosbag system~\cite{rosbag}. This system enables us to record messages published on specific topics and store them in a file, allowing us to reproduce them at will as if they were occurring in real-time. Within these rosbags, we capture messages containing map data, TF information, robot odometry, and sensor data. This approach ensures that we can evaluate different algorithms and configurations using identical input data, providing a reliable basis for comparison.

In our experimental setup, the robot is consistently teleoperated. We conduct various routes that commence within the interior of the structure, traverse both ascending and descending paths, and eventually transition outdoors, climbing slopes and descending into shallow depressions. The map had been previously created using our approach, in \cite{gazebo_gridmap_plugin} to generate the pair octomap/gridmap. 

\begin{figure}[h!]
  \centering
  \includegraphics[width=\linewidth]{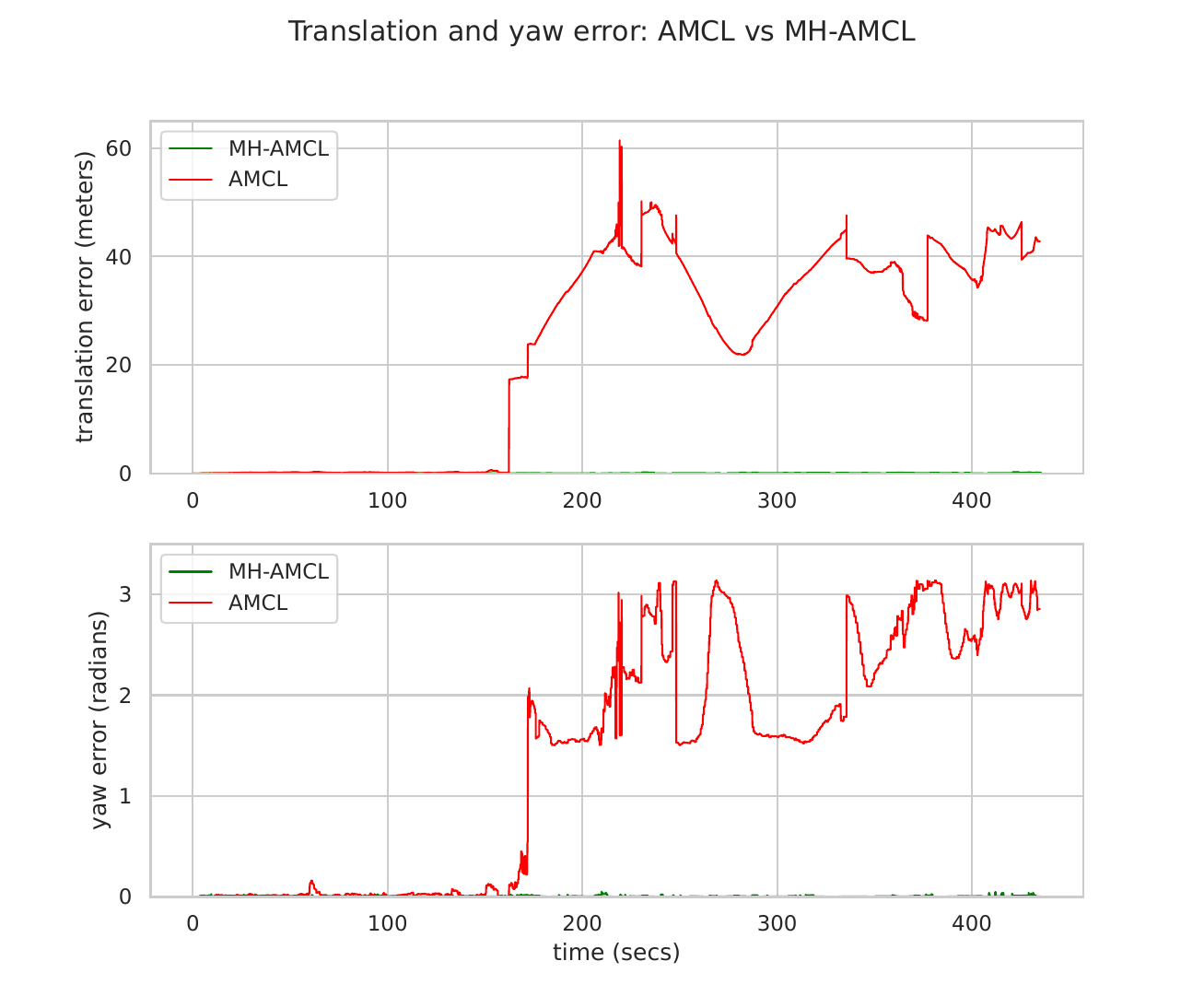}
\caption{Error comparative using AMCL and MH-AMCL in a non-planar simulated scenario with a robot using a 2D lidar. Errors are very low, near 0 for translation, for both approaches while the robot is on a plain floor, but AMCL completely fails as soon as the robot enters the house.}

  \label{fig:sim_amcl_vs_mh_amcl}
\end{figure}

In the first experiment, we compare the novel MH-AMCL algorithm presented in this paper with the well-known Nav2 AMCL. We will only use the 2D laser, as it is the only one that supports AMCL, to have a fair comparison. As depicted in Figure \ref{fig:sim_amcl_vs_mh_amcl}, the Nav2 AMCL algorithm struggles to accurately determine the robot's location. This issue arises due to changes in the laser inclination, making it challenging for the algorithm to align with the expected map. Consequently, this algorithm designed for flat scenarios is unsuitable for environments that force the robot to be inclined, with surfaces exhibiting varying elevations and terrain irregularities. In contrast, the MH-AMCL algorithm is capable of maintaining the robot's accurate position, with an average error of 0.087 m. in translation and 0.0047 radians in the orientation (only yaw angle, that we consider the critical component), being the maximum values of 0.28 m. and 0.0538 radians respectively.

In the second experiment, once the inability of AMCL in this scenario was shown, we focused on the properties of MH-AMCL with different configurations.
While we will present results from individual runs, the tables provided at the conclusion of this subsection consolidate findings from a total of 10 distinct routes, each lasting approximately 4 minutes.

\begin{figure}[h!]
  \centering
  \includegraphics[width=1.1\linewidth]{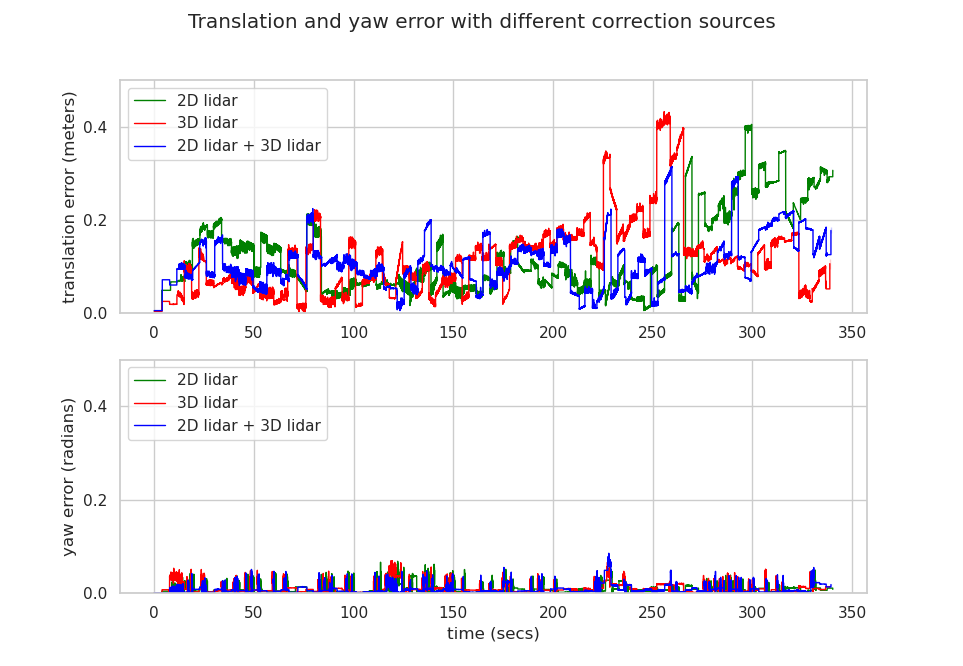}
\caption{Error comparative in a non-planar simulated scenario with different correction sources}
  \label{fig:sim_mh_amcl_error}
\end{figure}

Figure \ref{fig:sim_mh_amcl_error} shows the errors in estimating the robot's position in MH-AMCL using different configurations: only laser 2D, only laser 3D, and a fusion of both. Table \ref{tab:sensors} shows the average and maximum errors obtained using different configurations. As anticipated, employing a 2D sensor in real 3D environments shows the best results. While the outcomes are favorable, expanding the range with a 3D sensor leads to reductions in both maximum and average errors. The lowest error is achieved when both sensors are used in combination. Regardless, the error consistently remains below 0.4 m., predominantly hovering around 0.2 m., accompanied by negligible orientation errors. These results affirm the robust localization capabilities of the robot throughout the experiments.

\begin{table}[h!]
\centering
\caption{Summary of the mean and maximum errors (meters and radians) with different configurations in the simulation experiments.}
\label{tab:sensors}
\begin{tabular}{r|cc|cc|}
\cline{2-5}
\multicolumn{1}{c|}{}                     & \multicolumn{2}{c|}{Mean}             & \multicolumn{2}{c|}{Maximum}             \\ \cline{2-5} 
\multicolumn{1}{c|}{}                     & \multicolumn{1}{c|}{Translation} & Yaw   & \multicolumn{1}{c|}{Translation} & Yaw   \\ \hline
\multicolumn{1}{|r|}{2D lidar}            & \multicolumn{1}{c|}{0.123}       & 0.060 & \multicolumn{1}{c|}{0.404}       & 0.070 \\ \hline
\multicolumn{1}{|r|}{3D lidar}            & \multicolumn{1}{c|}{0.116}       & 0.008 & \multicolumn{1}{c|}{0.433}       & 0.073 \\ \hline
\multicolumn{1}{|r|}{2D lidar + 3D lidar} & \multicolumn{1}{c|}{0.106}       & 0.006 & \multicolumn{1}{c|}{0.314}       & 0.084 \\ \hline
\end{tabular}
\end{table}

\begin{figure}[h!]
  \centering
  \includegraphics[width=0.8\linewidth]{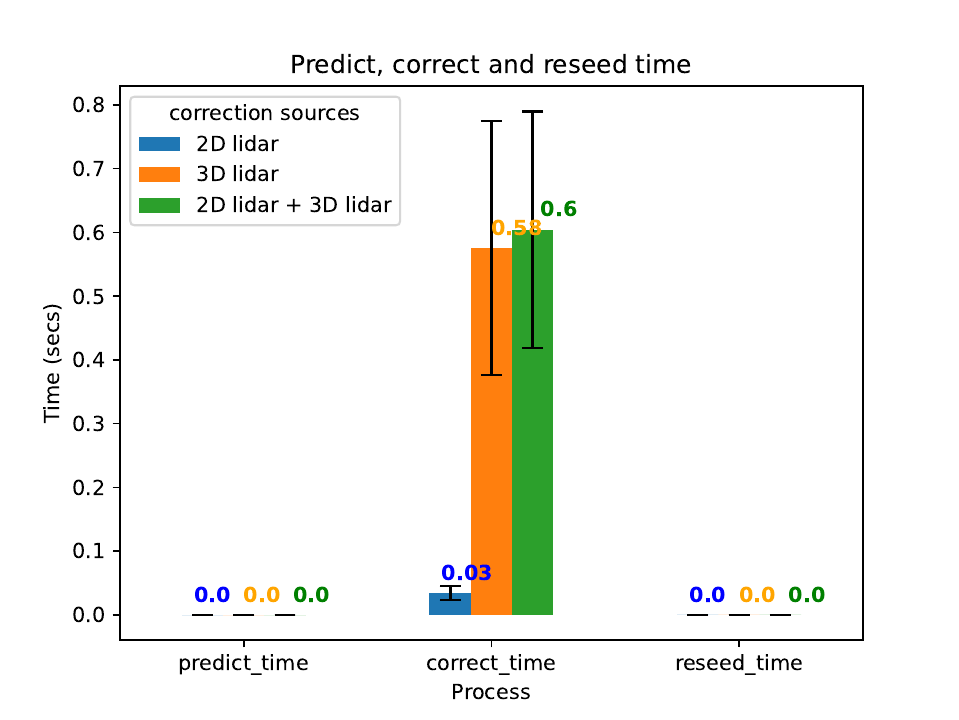}
\caption{Processing times in a non-planar simulated scenario with different correction sources}
  \label{fig:sim_mh_amcl_times}
\end{figure}

When considering processing times, we have considered the three stages comprising the localization process: prediction, correction, and reseed. In Figure \ref{fig:sim_mh_amcl_times}, we can observe the average values and their deviations for each utilized combination: 2D lidar, 3D lidar, and the combination of both. It is evident that the correction stage consumes the most time as it is responsible for executing the necessary movements to locate the robot accurately. In the case of the combination of 2D and 3D lidar, the average processing time is the sum of both. 

Videos of the execution can be found in \cite{video1} for the 3D sensor, \cite{video2} for the 2D sensor, and \cite{dataset} for the video playing the dataset used in the experiments. All the data and instructions to reproduce these results are in \cite{experiments}.

In the third experiment, we aim to validate the hypothesis that our quality value (Equation \ref{eq:mcl9}) offers a superior representation of a robot's correct localization compared to the analysis of uncertainty provided by the covariance matrix. To assess the covariance matrix, we will employ the "trace" metric, which involves the product of the main diagonal elements in the covariance matrix. The trace serves as a measure of the general uncertainty or variability in the state estimation. 

This test was conducted by manually relocating the robot to a new incorrect position 1-2 meters away from the correct position.  In Figure \ref{fig:recovery}, this results in a sharp drop in quality and slightly increased uncertainty. This demonstrates that the suggested quality metric can promptly detect when the robot is no longer accurately localized, whereas the covariance matrix requires more time to reach a state of awareness regarding this situation. As the robot decreases its error, the quality metric reflects this improvement.

\begin{figure}[h!]
  \centering
  \includegraphics[width=\linewidth]{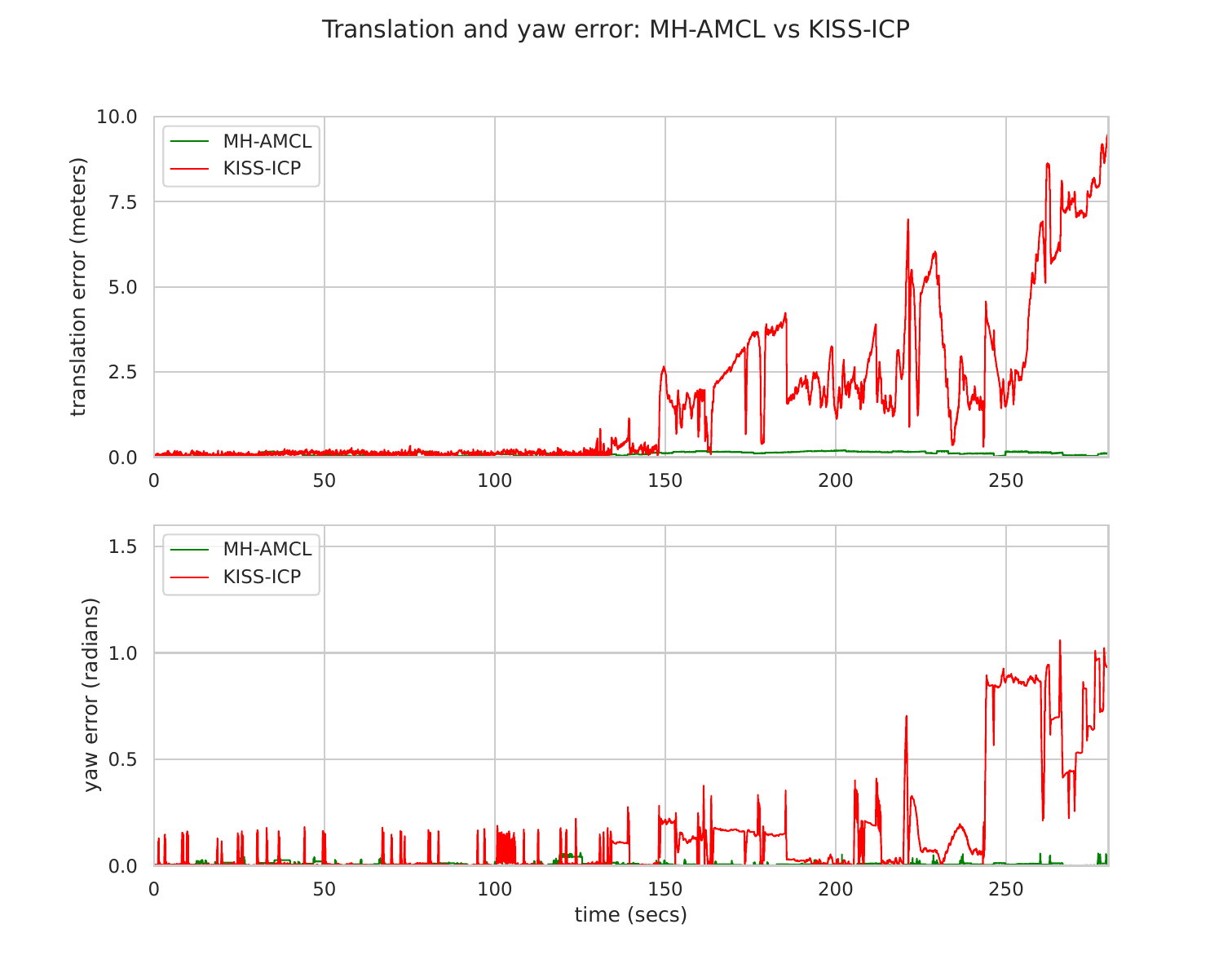}
\caption{Comparative analysis of KISS-ICP and MH-AMCL in a non-Planar simulated environment utilizing a robot equipped with 3D LiDAR technology. Errors remain low in both methods, but soon, KISS-ICP fails when the environment does not have much texture to make this approach work correctly.}
  \label{fig:kissicp}
\end{figure}

\begin{figure}[h!]
  \centering
  \includegraphics[width=\linewidth]{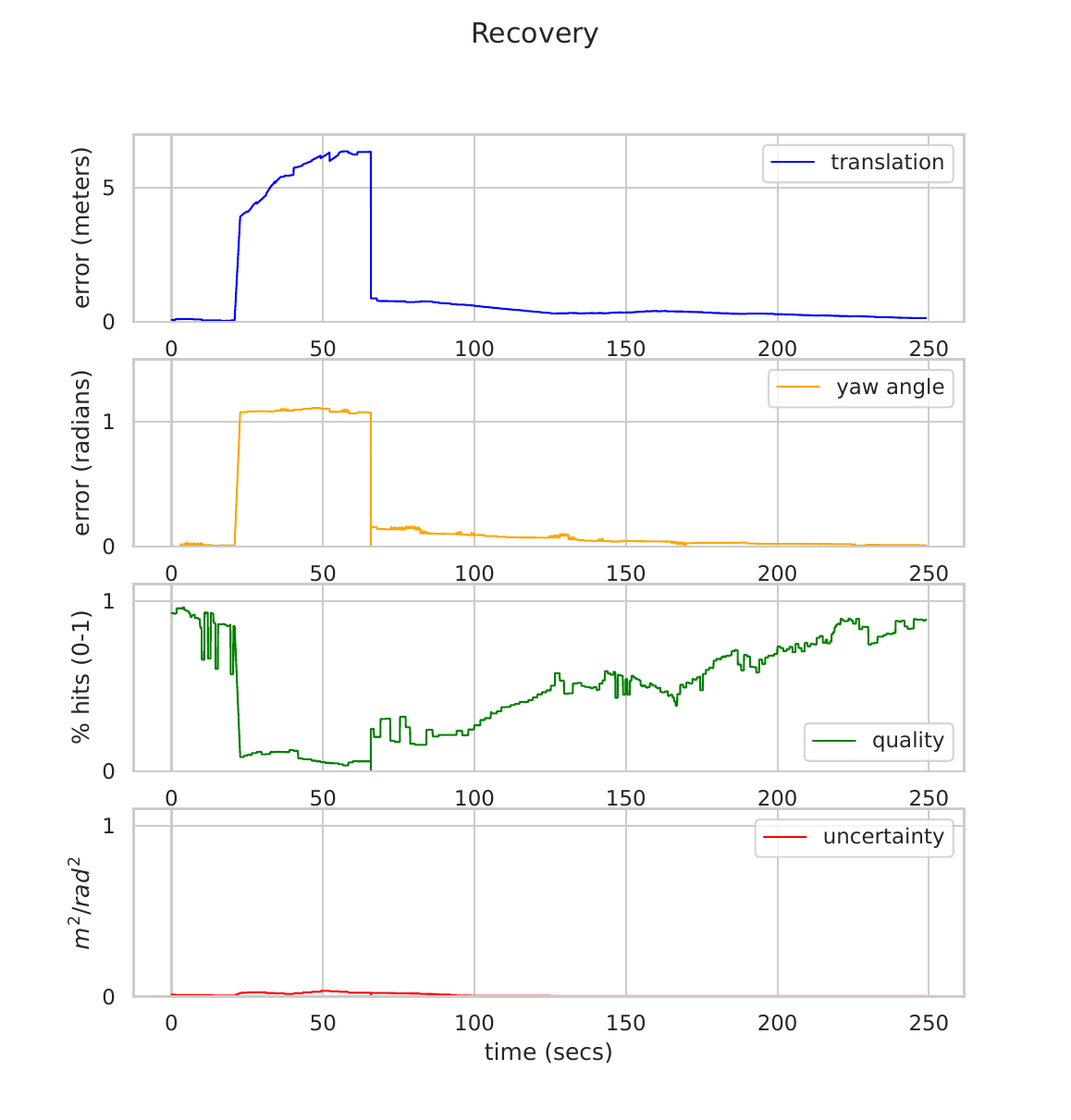}
\caption{Recovery in a non-planar simulated scenario with 2D and 3D lidar. The green line is the quality (Equation \ref{eq:mcl9}), and the red line is the covariance, calculated as the "trace" of the covariance matrix.}
  \label{fig:recovery}
\end{figure}

In our fourth experimental setup, the focus shifts to a comparative analysis between our proposal and the established KISS-ICP SLAM6D system~\cite{Vizzo20231029}, employing a 3D laser scanner in the simulated scenario. In this experiment, the robot starts from inside a building and extends to the outdoor surroundings, encompassing a complete circumnavigation of the structure. The results until the error of the SLAM 6D algorithm is below 10 meters are shown in Figure \ref{fig:kissicp}. Notably, when the robot exits the building and enters areas with reduced discriminative information, approximately at t=150, the 6D SLAM algorithm experiences a degradation in performance, resulting in inaccurate pose estimations and a lack of recovery capability. These open spaces, characterized by minimal surface textures, pose significant challenges for traditional SLAM 6D algorithms. Conversely, our proposed approach consistently maintains accurate estimations of the robot's position throughout this scenario. Table \ref{tab:kiss_icp_mh-amcl} summarizes the results for the whole experiment.


\begin{table}[]
\centering
\caption{Average and maximum errors: KISS-ICP vs MH-AMCL}
\label{tab:kiss_icp_mh-amcl}
\begin{tabular}{c|cc|cc|}
\cline{2-5}
                              & \multicolumn{2}{c|}{Average}             & \multicolumn{2}{c|}{Maximum}             \\ \cline{2-5} 
                              & \multicolumn{1}{c|}{Translation} & Yaw   & \multicolumn{1}{c|}{Translation} & Yaw  \\ \hline
\multicolumn{1}{|c|}{KISS-ICP}    & \multicolumn{1}{c|}{3.8397}       & 0.2095 & \multicolumn{1}{c|}{24.3198}       & 1.5686 \\ \hline
\multicolumn{1}{|c|}{MH-AMCL} & \multicolumn{1}{c|}{0.0951}       & 0.0076 & \multicolumn{1}{c|}{0.2476}       & 0.0734 \\ \hline
\end{tabular}
\end{table}

\subsection{Experiments at indoor}

We conducted a comprehensive real-world experiment to assess the effectiveness of our approach, employing a professional-grade robot, the Tiago Robot. To facilitate accurate ground truth measurements, we equipped the robot with a 2D laser, to which we affixed markers. This configuration allowed us to precisely determine the robot's actual position and orientation using the MOCAP4ROS2 system.

The mapping process consists of a few steps. First, get the point cloud representing the restructure of the environment using the package lidarslam\_ros2 \cite{lidarslam_ros2}. Next, we create the pair octomap/gridmap from this point cloud using the package gridmap\_slam \cite{gridmap_slam}. Once generated and saved to disk, this pair octomap/gridmap is published in each execution using the package extended\_map\_server \cite{mapserver}.

Similar to the previous section, we conducted 10 test runs, each lasting approximately 2 minutes. These experiments were meticulously recorded using rosbags, enabling us to apply various configurations of our localization algorithms for evaluation.

\begin{figure}[h!]
  \centering
  \includegraphics[width=\linewidth]{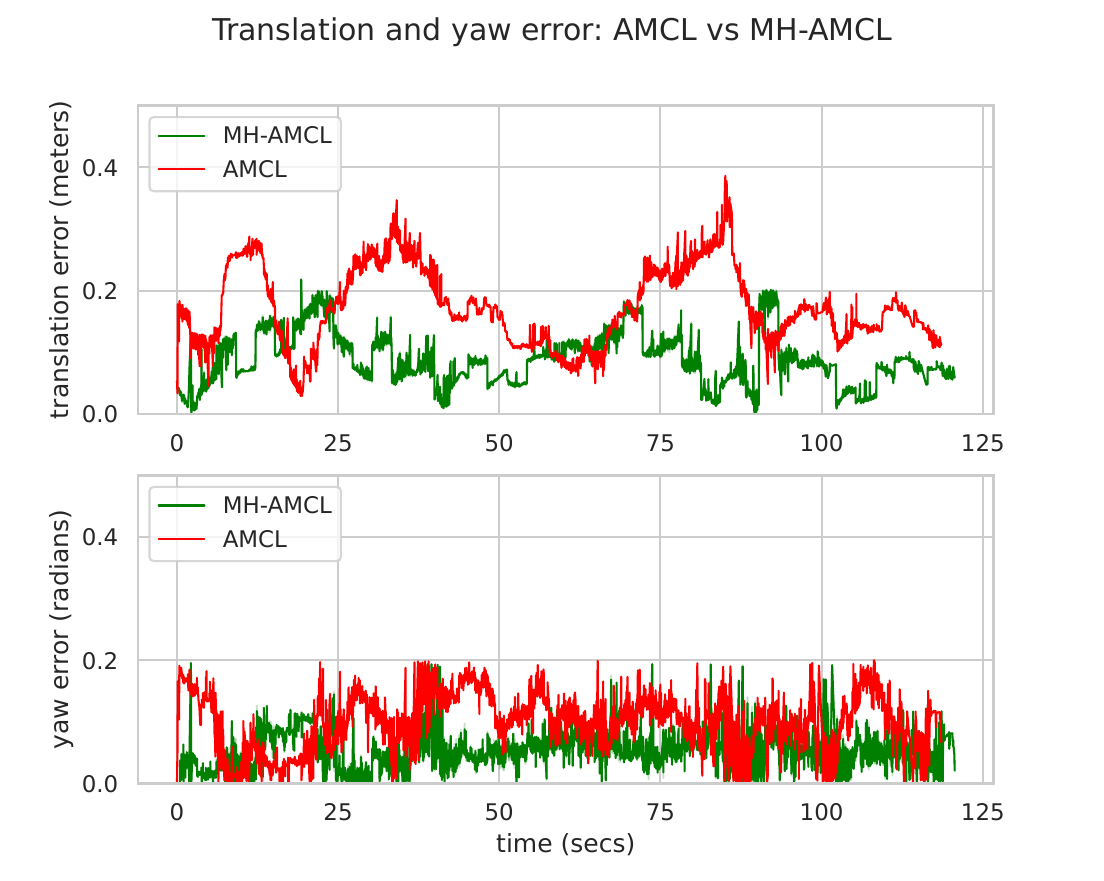}
\caption{Comparative using AMCL and MH-AMCL in a real planar scenario with a robot using a 2D lidar}
  \label{fig:real_amcl_vs_mh_amcl}
\end{figure}

This experiment entails a comparative analysis between MH-AMCL and the reference nav2 AMCL implementation. Our assessment focuses on the precision of these algorithms, specifically evaluating the errors in position and orientation estimation. The detailed results are presented in Figure \ref{fig:real_amcl_vs_mh_amcl}, and a comprehensive summary of all the attempts is provided in Table \ref{tab:real_amcl_mh-amcl}. The outcomes obtained with both algorithms exhibit striking similarities, demonstrating their precision and reliability, with errors consistently below 0.4 meters and 0.2 radians, which are considered good results. It is important to note that AMCL (this implementation, in particular) is widely recognized as the state-of-the-art solution for indoor robot localization and has been deployed across thousands of robots in various industrial applications over the past two decades. As highlighted in the preceding section, AMCL's limitations become apparent when confronted with non-flat environments. However, through our experimentation, we have demonstrated that our proposed approach yields performance on par with, if not superior to, AMCL in such scenarios.

\begin{table}[]
\centering
\caption{Average and maximum errors (meters and radians): AMCL vs MH-AMCL}
\label{tab:real_amcl_mh-amcl}
\begin{tabular}{c|cc|cc|}
\cline{2-5}
                              & \multicolumn{2}{c|}{Average}             & \multicolumn{2}{c|}{Maximum}             \\ \cline{2-5} 
                              & \multicolumn{1}{c|}{Translation} & Yaw   & \multicolumn{1}{c|}{Translation} & Yaw  \\ \hline
\multicolumn{1}{|r|}{AMCL}    & \multicolumn{1}{c|}{0.172}       & 0.099 & \multicolumn{1}{c|}{0.386}       & 0.199 \\ \hline
\multicolumn{1}{|r|}{MH-AMCL} & \multicolumn{1}{c|}{0.086}       & 0.05  & \multicolumn{1}{c|}{0.219}       & 0.195 \\ \hline
\end{tabular}
\end{table}

\subsection{Experiments with real robot at indoor/outdoor}

As we presented before, the last experiment was conducted using the Summit XL robot in a teleoperated itinerary outside and inside the laboratory building, traversing irregular terrain in the pathways, including ramps in certain areas. The mapping process is similar to the one used in the previous experiment.

\begin{table}[]
\centering
\caption{Average and standard deviation error in meters: SLAM 3D vs MH-AMCL}
\label{tab:lidarslam_mh-amcl}
\begin{tabular}{|l|c|c|}
\hline
\multicolumn{1}{|c|}{} & MH-AMCL & SLAM 3D \\ \hline
Trial 0                & $0.738 \pm 0.551$ & $1.214 \pm 0.503$ \\ \hline
Trial 1                & $0.600 \pm 0.457$ & $1.321 \pm 0.714$ \\ \hline
Trial 2                & $0.629 \pm 0.321$ & $0.993 \pm 0.424$ \\ \hline
Trial 3                & $0.555 \pm 0.352$ & $2.113 \pm 1.058$ \\ \hline
Trial 4                & $0.540 \pm 0.282$ & $1.178 \pm 0.493$ \\ \hline
Trial 5                & $0.862 \pm 0.702$ & $2.043 \pm 0.913$ \\ \hline
Trial 6                & $0.521 \pm 0.297$ & $1.172 \pm 0.546$ \\ \hline
Trial 7                & $0.607 \pm 0.294$ & $1.280 \pm 0.507$ \\ \hline
\end{tabular}
\end{table}

\begin{figure*}[bh!]
  \centering
  \includegraphics[width=1\linewidth]{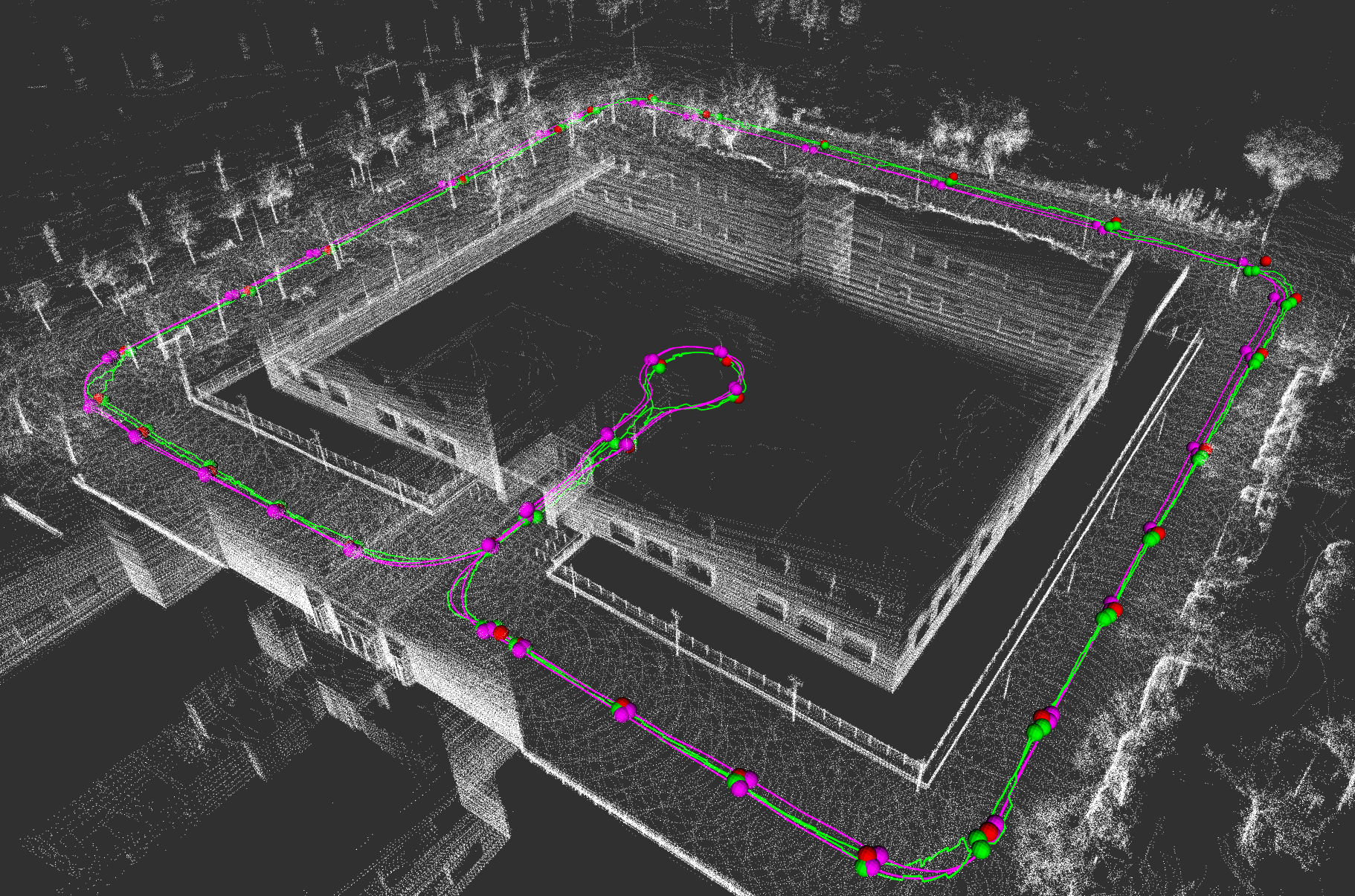}
  \caption{Itinerary of the robots in two trials per algorithm. Red spheres are ground truth control points. Green lines and control points are in green for MH-AMCL and fuchsia for SLAM-3D.}
  \label{fig:campus_exp}
\end{figure*}

This particular experiment was replicated seven times, with all the robot's sensory data captured in a rosbag file. This approach enables the subsequent playback of each trial for analysis using various algorithms and settings.

Given the absence of a ground truth system in this setting, we established a series of checkpoints along the navigation path. At these checkpoints, the precise positions were measured and recorded in map coordinates. The robot was directed to traverse these points, with the instances of it passing directly over each marker documented. These recordings are included in the rosbag files for future playback and analysis.

In our study, we chose not to benchmark against AMCL, our standard reference, due to its design for flat terrains and compatibility with robots that utilize only 2D laser sensors. It's important to note that adapting AMCL for use in three-dimensional environments represents one of the innovative aspects of our research. Consequently, we opted to compare our methodology against a contemporary 3D SLAM algorithm that stands at the forefront of current technological advancements. Specifically, our comparison targets a ROS 2 slam package that employs OpenMP-enhanced gicp/ndt scan matching for its frontend and a graph-based SLAM approach for its backend, as detailed in \cite{lidarslam_ros2}. While these SLAM algorithms are recognized for their precision, their lack of reliance on a predefined map often leads to challenges in achieving map closure, particularly over the extensive routes featured in our experiments. This analysis will also include running the algorithm on the rosbags recorded during our tours.

To assess the effectiveness of our approach, we will employ the Euclidean distance metric, comparing the robot's location as estimated by the algorithm against its actual, measured position at various control points.

\begin{figure}[h!]
  \centering
  \includegraphics[width=\linewidth]{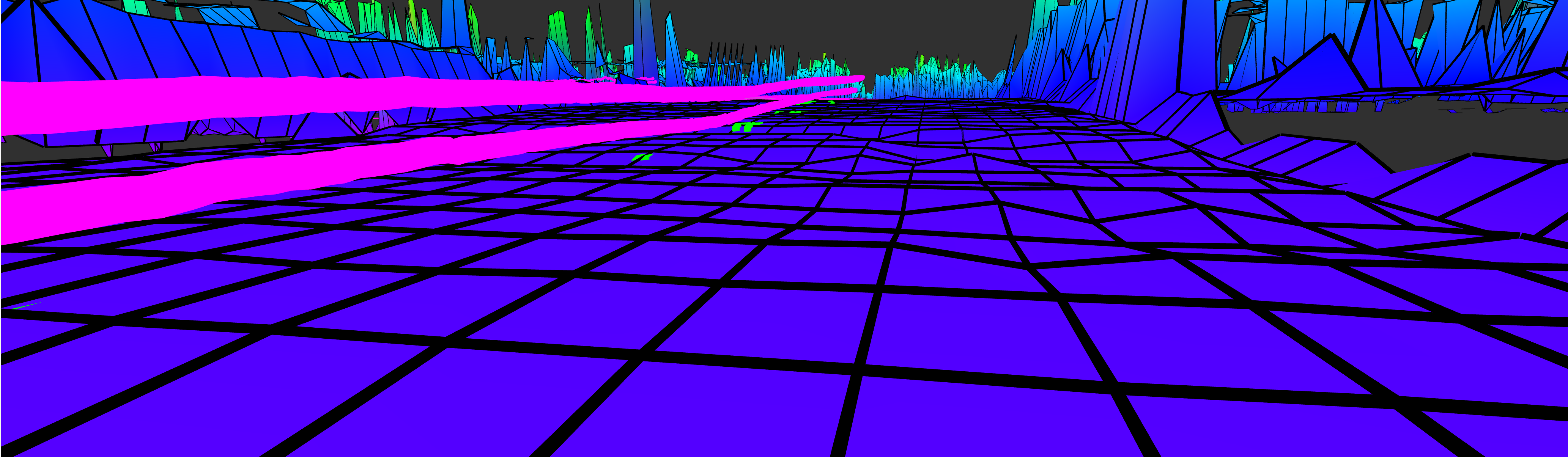}
\caption{In the last part of the trials, while the green path (MH-AMCL) is in the ground, the fuchsia path (SLAM 3D) is above the floor, not able to resolve the loop closure for the ground.}
  \label{fig:loop_closure}
\end{figure}

The Figure \ref{fig:campus_exp} illustrates the robot's entire trajectory in two different trials. Control points are denoted by red spheres, with paths generated by our MH-AMCL algorithm highlighted in green and those generated by the 3D SLAM algorithm depicted in fuchsia. The spheres represent the location estimates at each control point, color-coded in green for MH-AMCL and in fuchsia for 3D SLAM. Figure \ref{fig:loop_closure} specifically highlights the difficulties encountered by 3D SLAM algorithms in achieving loop closure on the ground across various tests.

The analytical data of the experiment is shown in Table \ref{tab:lidarslam_mh-amcl}. The findings demonstrate that our algorithm performs effectively, maintaining errors within an average of 0.9 meters. Despite achieving favorable outcomes, the 3D SLAM algorithm's error margin widens during the final part of the trial, attributed to inaccuracies in loop closure and cumulative mapping errors over the course. Based on these observations, we can conclude that our algorithm exhibits adequate precision and robustness in varied environments, both indoor and outdoor. It surpasses the performance of existing state-of-the-art algorithms, presenting a reliable alternative to solely relying on GPS, which is commonly acknowledged for its lesser accuracy.

\section{CONCLUSIONS}\label{sec:conclusions}

In this paper, we have introduced our mobile robot localization algorithm, which is based on AMCL and tailored for non-planar environments. Our approach leverages elevation maps (gridmaps) and probabilistic 3D occupancy maps (octomaps) to represent the robot's surroundings. Our method's key advantage is its ability to locate a robot in non-planar environments, often encountered in unstructured outdoor settings or indoor environments featuring ramps. In such scenarios, robots can be inclined and perceive obstacles absent in traditional 2D maps. Furthermore, our approach offers flexibility in integrating various types and quantities of range sensors while considering the unique characteristics of each sensor. These advancements lay the groundwork for comprehensive navigation in non-planar terrains.

Our approach has been seamlessly integrated into Nav2, the open-source standard for robot navigation within the ROS community. This integration not only validates our work's functionality and the robustness of the results demonstrated in the validation section but also makes it readily accessible for deployment by companies in their robotic systems and for use by researchers pushing the boundaries of knowledge in this field. Experimental trials have affirmed its effectiveness and precision in non-planar environments and with professional-grade robots.

We are currently expanding this work in several directions. Firstly, our implementation is poised to support multiple hypotheses regarding the robot's position, although we are actively developing a module to generate additional hypotheses for the robot's potential locations. This feature is invaluable when a robot is moved manually or starts in unknown positions. Another avenue of exploration is the full integration of elevation maps into navigation algorithms, enabling path planning that avoids gradients and enhances a robot's energy efficiency. Finally, the inclusion of terrain features holds promise for enhancing the accuracy of the robot's motion modeling. By integrating supplementary gridmap layers that explicitly represent these terrain characteristics, we can effectively incorporate them into the particle filter's prediction phase of the motion model or in the path planner in future works about navigation.

\section*{Author Contributions}
Conceived and developed, F.M.R., J.M.G, and R.P.R.; Conceived and designed the experiments, F.M.R and J.M.G.; Performed the experiments, J.M.G., J.D.P, and A.G.J.; All authors wrote the paper. All authors have read and agreed to the published version of the manuscript.

\section*{Acknowledgements}
This work is partially funded under Project PID2021-126592OB-C22 funded by MCIN/AEI/
10.13039/501100011033, the grant TED2021-132356B-I00 funded by MCIN/AEI/10.13039/501100011033 and by the “European Union NextGenerationEU/PRTR", and by CORESENSE project with funding from the European Union’s Horizon Europe Research and Innovation Programme (Grant Agreement No. 101070254). The authors also would like to thank Juan Carlos Manzanares, who helped us to carry out the experiments.

\bibliography{references}%

\end{document}